\documentclass{article}

\newcommand{\bs}[1]{\boldsymbol{#1}}

\usepackage{graphicx}
\usepackage{amsmath}
\usepackage{amsthm}
\usepackage{amssymb}
\usepackage{mathtools}
\usepackage{xcolor}

\newcommand{\shrink}[1]{}

\newtheorem{remark}{Remark}[section]
\newtheorem{theorem}{Theorem}[section]

\newtheorem{lemma}{Lemma}[section]
\newtheorem{algorithm}{Algorithm}[section]
\newtheorem{assumption}{Assumption}[section]

\usepackage{times}
\usepackage{graphicx} 
\usepackage{subfigure}

\usepackage{natbib}

\usepackage{algorithm}
\usepackage{algorithmic}

\usepackage{hyperref}

\usepackage[accepted]{icml2019}

\usepackage{times}
\usepackage{xcolor}
\usepackage{soul}
\usepackage[utf8]{inputenc}

\usepackage{latexsym}
\usepackage{booktabs}
\usepackage{amsmath}
\usepackage{verbatim}
\usepackage{amssymb}
\usepackage{amsthm}
\usepackage{algorithm}
\usepackage{algorithmic}
\usepackage{graphicx}
\usepackage{subfigure}
\usepackage{natbib}
\usepackage{booktabs}
\usepackage{epstopdf}
\usepackage{lipsum}

\usepackage[utf8]{inputenc} 
\usepackage[T1]{fontenc}    
\usepackage{url}            
\usepackage{booktabs}       
\usepackage{amsfonts}       
\usepackage{nicefrac}       
\usepackage{microtype}      

\usepackage{multicol}

\usepackage{url}

\usepackage{wrapfig}

\icmltitlerunning{Online Alternating Minimization}

\begin{document}

\twocolumn[
\icmltitle{Beyond Backprop:    Online  Alternating Minimization with Auxiliary Variables}









 
\icmlsetsymbol{equal}{*}

\begin{icmlauthorlist}
\icmlauthor{Anna Choromanska*}{NYU}
\icmlauthor{Benjamin Cowen*}{NYU}
\icmlauthor{Sadhana Kumaravel*}{IBM}
\icmlauthor{Ronny Luss*}{IBM}
\icmlauthor{Mattia Rigotti*}{IBM}
\icmlauthor{Irina Rish*}{IBM}
\icmlauthor{Brian Kingsbury}{IBM}
\icmlauthor{Paolo DiAchille}{IBM}
\icmlauthor{Viatcheslav Gurev}{IBM}
\icmlauthor{Ravi Tejwani}{MIT}
\icmlauthor{Djallel Bouneffouf}{IBM}

\end{icmlauthorlist}

  \icmlcorrespondingauthor{Irina Rish}{IBM}

\icmlaffiliation{NYU}{ ECE NYU Tandon}
\icmlaffiliation{IBM}{IBM T.J. Watson Research Center}
\icmlaffiliation{MIT}{MIT}

\icmlkeywords{deep networks, optimization, online algorithms, BCD alternating minimization}

\vskip 0.3in
]


\printAffiliationsAndNotice{\icmlEqualContribution} 

\begin{abstract}

Despite significant recent advances in deep neural networks,  training them remains a challenge due to the highly non-convex nature of the objective function.  State-of-the-art methods rely on error backpropagation, which suffers from   several well-known issues, such as vanishing and exploding gradients, inability to handle non-differentiable nonlinearities and to parallelize weight-updates across layers, and biological implausibility. These limitations continue to motivate exploration of alternative training algorithms,   including several recently proposed auxiliary-variable methods  which break the complex nested objective function into local subproblems. However, those techniques are mainly offline (batch), which limits their applicability to   extremely large datasets, as well as to online, continual or reinforcement learning.  The main contribution of our work is  a    novel online (stochastic/mini-batch) alternating minimization (AM) approach  for training deep neural networks, together with the first theoretical convergence guarantees for AM in stochastic settings and promising empirical results  on a variety of architectures  and datasets.
\end{abstract}

\vspace{-0.2in}
\section{Introduction}
\vspace{-0.05in}
\label{sec:intro}
 
 
Backpropagation (backprop)  \cite{rumelhart1986learning}  has been the workhorse of neural net learning for several decades, and its practical effectiveness 
is demonstrated by recent successes of deep learning in a wide range of applications. Backprop (chain rule differentiation) is used to compute gradients in state-of-the-art learning algorithms such as
stochastic gradient descent (SGD) \cite{robbins1985stochastic} and its variations \cite{duchi2011adaptive,tieleman2012lecture,zeiler2012adadelta,kingma2014adam}.

However, backprop has several drawbacks as well, including the commonly known   {\em vanishing gradient} issue, resulting from
recursive application of the chain rule through multiple layers of  deep and/or recurrent networks \cite{bengio1994learning,riedmiller1993direct,hochreiter1997long,pascanu2013difficulty,goodfellow2016deep}.  Although several approaches were proposed to address this issue, including Long Short-Term Memory \cite{hochreiter1997long}, RPROP \cite{riedmiller1993direct}, and rectified linear units (ReLU) \cite{nair2010rectified}, the fundamental problem with computing gradients of a deeply nested objective function remains. 
Moreover,  backpropagation does not apply directly to {\em non-differentiable nonlinearities} and  {\em does not allow parallel weight updates}  across the   layers  \cite{le2011optimization,carreira2014distributed,taylor2016training}.  

Also, besides its computational issues,    backprop is  often criticized from a neuroscience perspective as a biologically implausible learning mechanism \cite{lee2015difference,bartunov2018assessing,krotov2019unsupervised,sacramento2018dendritic,guerguiev2017towards}, due to multiple factors including the need for  "a distinct form of information propagation (error feedback) that {\em does not influence neural activity}, and hence does not conform to known biological feedback mechanisms underlying neural communication" \cite{bartunov2018assessing}\footnote{Gradient chain computation yields   {\em non-local}  synaptic weight updates which depend on the activity and computations of all downstream neurons, rather than only local signals from adjacent neurons  \cite{whittington2019theories,krotov2019unsupervised}.}.

The issues mentioned above continue to motivate research on alternative algorithms for neural net learning.  Several approaches were proposed recently, introducing {\em auxiliary variables} associated with hidden unit activations in order to decompose the  highly coupled problem of optimizing a  nested loss function  into  multiple, loosely coupled, simpler subproblems.
 These include alternating direction method of multipliers (ADMM) \cite{taylor2016training,zhang2016efficient} and alternating-minimization or block coordinate descent (BCD) methods \cite{carreira2014distributed,zhang2017convergent,Zhang2017,Askari2018,zeng2018block,lau2018proximal,gotmare2018decoupling}. 
 
 A similar formulation, using Lagrange multipliers,   was proposed  earlier in \cite{le1986learning,yann1987modeles,lecun1988theoretical}, where a constrained formulation involving activations required the output of the previous layer to be equal to the input of the next layer, leading to the    {\em target propagation} algorithm and recent extensions \cite{lee2015difference,bartunov2018assessing} (unlike BCD and ADMM, target prop uses layer-wise inverses of the forward mappings). These methods  are viewed as somewhat more  bio-plausible alternatives to backprop due to explicit  propagation of (noisy/nondeterministic)  neuronal activity and (layer-)local synaptic updates  (see \cite{bartunov2018assessing} for details). Note that the above bio-plausibility arguments are equally applicable   to auxiliary-variable methods based on explicit optimization of (noisy) neural activations, and   breaking the weight update problem into local, layer-wise optimization subproblems. 


In this paper, we propose a {\em novel  activation-propagation approach}, which,  similarly to prior BCD and ADMM approaches, performs alternating   minimization of network weights and auxiliary activation variables. However, unlike those methods,  which all assume an offline (batch) setting and  require the full training dataset at each iteration, our method is   an {\em online}, incremental learning approach, that performs {\em stochastic (minibatch) alternating minimization (AM)}. Two  variants of AM are proposed, {\em AM-Adam} and {\em AM-mem}, which use different approaches for optimizing local subproblems.

Note that, unlike ADMM-based methods \cite{taylor2016training,zhang2016efficient} 
  and some previously proposed  BCD methods  \cite{zeng2018block}, our  approach  does not require Lagrange multipliers and only uses one set of auxiliary variables per layer: it is  as memory-efficient as standard SGD, which stores activation values for gradient computations. The same distinction, along with  multiple others (discussed in Supplementary Material), exists between our method and another recently proposed alternating-minimization scheme, ProxProp \cite{Frerix-et-al-18}. Also, we assume arbitrary  loss functions  and nonlinearities (unlike, for example, \cite{zhang2017convergent} which assumes ReLU nonlinearities), and perform extensive empirical evaluation beyond the fully-connected networks, commonly used to evaluate  auxiliary-variable methods.
 
 In summary, our contributions include:\\
 \vspace{-0.3in}
\begin{itemize}
\item {\em algorithm(s):} a novel {\em online (mini-batch) auxiliary-variable approach} for training neural networks without the gradient chain rule of backprop; unlike prior  offline (batch) auxiliary-variable algorithms, our method  can scale to arbitrarily large datasets and is   applicable in continual and reinforcement learning settings;
 \vspace{-0.1in}
\item {\em theory:} to the best of our knowledge, we propose the first general theoretical convergence guarantees of alternating minimization in the stochastic setting.  We show that the error of AM decays at the sub-linear rate $O((1/t)^{3/2} + 1/t)$ as a function of the iteration  $t$;
 \vspace{-0.1in}
\item {\em extensive empirical evaluation} on a variety of network architectures and datasets, demonstrating significant advantages of our method vs. offline counterparts, as well as somewhat faster initial convergence as compared to SGD and Adam, followed by similar asymptotic performance;
 \vspace{-0.1in}
\item our online method inherits common advantages  of  similar offline auxiliary-variable methods, including (1) {\em no vanishing  gradients}, (2) handling of {\em non-differentiable nonlinearities} more easily in local subproblems, and 
(3) the {\em possibility for  parallelizing weight updates across  layers};
 \vspace{-0.1in}
\item  similarly to target propagation approaches \cite{le1986learning,yann1987modeles,lee2015difference,bartunov2018assessing}, our method is based on an {\em explicit propagation of neural activity} and {\em local synaptic updates}, which is one step closer to a more {\em biologically plausible} credit assignment mechanism than backprop; see   \cite{bartunov2018assessing} for a  detailed discussion on this topic.
  \end{itemize}



\vspace{-0.1in}
\section{Alternating Minimization: Breaking Gradient Chains with Auxiliary Variables}
\vspace{-0.05in}
\label{sec:method}
We denote as  $(\bs{X}, \bs{Y})=$ $\{(\bs{x}_1,\bs{y}_1),...,(\bs{x}_n, \bs{y}_n) \}$ a dataset of $n$ labeled samples, where $\bs{x}_t$ and $\bs{y}_t$ are the   sample and its (vector) label at time $t$, respectively (e.g.,  one-hot  $m$-dimensional vector  $\bs{y}$ encoding  discrete labels with  $m$ possible values).  We assume   $\bs{x} \in \mathbb{R}^N$, and  $\bs{y} \in \{0,1\}^m$.
Given a fully-connected neural network with $L$ hidden layers,  
$\bs {W^j}$  denotes the $m_j \times m_{j-1}$ link weight matrix associated with the links from layer $j-1$ to layer $j$,  where $m_j$ is the number of nodes at layer $j$. $\bs{W}^{L+1}$ denotes the  $m_L \times m$ weight matrix connecting the last hidden layer $L$ with the output. We denote the set of all weights
$\bs{W}=\{\bs{W}^1,...,\bs{W}^{L+1}\}$.

\noindent{\bf Optimization problem.} Training a fully-connected neural network with $L$ hidden layers   consists of minimizing, with respect to weights  ${\bs W}$, the loss
${\cal{L}}(y,f({\bs W},\bs{x}_L))$ involving a nested function  $f({\bs W},\bs{x}_L)=f_{L+1}(\bs{W}_{L+1},f_L(\bs{W}_L,f_{L-1}(\bs{W}_{L-1},...f_1(\bs{W}_1,\bs{x})...)$; this can be re-written as constrained optimization:
\begin{eqnarray}
    \nonumber 
\min_{\bs{W}} & \sum_{t=1}^n {\cal{L}}(\bs{y}_t,\bs{a}_t^{L},\bs{W}^{L+1}),~where~ \bs{a}_t^l = \sigma_{l}(\bs{c}^l_t), 
 \\  & s.t. ~ \bs{c}^l_t=\bs{W}^l \bs{a}_t^{l-1},~ l=1,...,L, ~and ~~  \bs{a_t}^0=\bs{x}_t.
 \label{eq:NN}
\end{eqnarray}
In the above formulation, we use $\bs{a}_t^l$ as shorthand (not a new variable)   denoting  the {\em activation} vector of hidden units in layer $l$,  where $\sigma$ is a nonlinear activation function (e.g, ReLU, $tanh$, etc) applied to {\em code} $\bs{c}^l$, a new {\em auxiliary variable} that must be equal to a linear transformation of the previous-layer activations.  
 
For classification problems, we use the multinomial loss as our objective function: \quad${\cal{L}}(\bs{y}, \bs{x},\bs{W})=-\log{P(\bs{y} | \bs{x}, \bs{W})}$
\vspace{-0.05in}
\begin{equation}
\begin{array}{ll}
=& -\displaystyle\sum_{i=1}^{m}{\bs{y}_i(\bs{w}_i^T \bs{x})}+ \log{(\displaystyle\sum_{l=1}^{m}{\exp{(\bs{w}_l^T\bs{x})}})},
\end{array}
\vspace{-0.02in}
\end{equation}
where $w_i$ is the $i^{th}$ column of $\bs{W}$, $y_i$ is the $i^{th}$ entry of the one-hot vector encoding $\bs{y}$,  and the class likelihood is modeled as $P(y_i=1 | \bs{x}, \bs{W}) =$ $\exp{(\bs{w}_i^T\bs{x})}/\sum_{l=1}^{m}{\exp{(\bs{w}_l^T\bs{x})}}$.

\shrink{
\cite{carreira2014distributed} and \cite{taylor2016training}, we use alternating minimization. However, we develop an online approach while both previous approaches are formulated in an offline batch mode that learns from a whole training dataset rather than incrementally. Such approaches have limited scalability to extremely large  datasets (even using  parallelization across samples as in \cite{taylor2016training}, and, more importantly, cannot handle  online, continual learning scenarios, unlike the standard backpropagation-based stochastic gradient methods. }

\noindent{\bf Offline Alternating Minimization.} We start with an offline optimization problem formulation,  for a given dataset of $n$ samples, which is similar to   \cite{carreira2014distributed} but uses multinomial instead of  quadratic loss, and a different set of {\em auxiliary variables}. Namely,  we use the following relaxation of the constrained formulation in eq. \ref{eq:NN}:
\vspace{-0.1in}
\begin{equation}
\begin{split}
 f(\bs{W},\bs{C}) & = \sum_{t=1}^n {\cal{L}}(y_t, \sigma_L(\bs{c}_t^L), \bs{W}^{L+1})\\
& \quad + \mu \sum_{t=1}^n \sum_{l=1}^{L}  ||\bs{c}_t^l - \bs{W}^l \sigma_{l-1}(\bs{c}_t^{l-1}) ||_2^2.
\end{split}
 \label{obj1}
\end{equation}
This problem can be   solved by alternating minimization (AM), or block-coordinate descent (BCD), over weights $\bs{W}=\{\bs{W}^1,...,\bs{W}^{L+1}\}$ and codes $\bs{C}=\{\bs{c}_1^1,...,\bs{c}_1^L,$ $...\bs{c}_n^1,...,\bs{c}_n^L,\}$. Each iteration involves  optimizing  $\bs{W}$ for fixed $\bs{C}$, followed by fixing $\bs{W}$ and   optimizing $\bs{C}$.
The parameter $\mu > 0$ acts as a regularization weight.
As in \cite{carreira2014distributed}, we use an adaptive scheme for gradually increasing $\mu$ over iterations\footnote{  
 Note that sparsity ($l_1$  regularization) on both $\bs{c}$ and $\bs{W}$  could be easily  added to the objective in eq. \ref{obj1} and would not change the computational complexity of the algorithms detailed below (we can use proximal instead of  gradient methods).}



    \noindent{\bf Online Alternating Minimization.} The offline alternating minimization outlined above is not scalable to extremely large datasets (even data-parallel methods, such as \cite{taylor2016training}, are inherently limited by the number of cores available), and   not suitable for incremental, continual/lifelong \cite{ring1994continual,thrun1995lifelong,thrun1998lifelong} or reinforcement learning scenarios with  potentially infinite data streams. 
To overcome those limitations, we propose a general {\em  online AM } algorithmic scheme and present {\em two specific algorithms}  which differ in   optimization approaches used for updating $\bs{W}$; both algorithms are later evaluated and compared empirically.

Our approach is outlined in Algorithms \ref{alg:AM}, \ref{alg:AM2}, and \ref{alg:AM3}, omitting implementation details such as the adaptive $\mu$ schedule, hyperparameters controlling the number of iterations in   optimization subroutines, and several others; we will make our code available online.
  As an input, the method takes an initial   $\bs{W}$ (e.g., random), initial penalty weight $\mu$, learning rate for the predictive layer, $\eta$, and a Boolean variable $Mem$, indicating which optimization method to use for $\bs{W}$ updates; if $Mem=1$, a memory-based approach  (discussed below) is selected, and  initial memory matrices $\bs{A_0}$, $\bs{B_0}$ (described below) will be provided (typically, both are  initialized to all zeros unless we want to retain the memory of some prior learning experience, e.g.  in a continual learning scenario).    The algorithm processes samples one at a time (but can easily be generalized to  mini-batches); the current sample  is encoded in its  representations at each layer ({\bf encodeInput} procedure, Algorithm \ref{alg:AM2}),  and an output prediction is made  based on such encodings. The prediction error is computed, and the backward code updates follow as shown in the {\bf updateCodes} procedure, where the code vector at layer $l$ is optimized with respect to the only two parts of the global objective that the code variables participate in.
  Once the codes are updated, the {\em weights can be  optimized in parallel across the layers} (in {\bf updateWeights} procedure, Algorithm \ref{alg:AM3}) since fixing codes breaks the weight optimization problem into layer-wise independent subproblems. We next discuss each step in detail.

\begin{small}
\begin{algorithm}[ht]
\caption{
 Online Alternating Minimization (AM) 
}
\label{alg:AM}
 \begin{algorithmic}[1]
{\small 
\REQUIRE $(\bs{x},\bs{y}) \sim p(\bs{x},y)$ (data stream sampled from distribution  $p(\bs{x},\bs{y})$; initial weights $\bs{W}_0$;   $\mu \in \mathbb{R^+}$ (quadratic penalty weight); $\eta \in \mathbb{R^+}$ (top-layer weight update step size); $Mem$ (indicates the type of optimization method for  {\bf updateWeights}; if "yes", input initial memory matrices $\bs{A_0}$ and $\bs{B_0}$).
 \vspace{0.05in}
\WHILE{more samples}
 \STATE Input $(\bs{x}_t, y_t)$ 
 \label{as:input_data}
\STATE $\bs{C} \gets$ {\bf encodeInput}($~\bs{x}_t$,$~\bs{W}_{t-1}$)  \% forward: compute linear activations at layers $1,...,L$
\STATE $\bs{C} \gets$ {\bf updateCodes}($~\bs{C}$,$~y_t$, $~\bs{W}_{t-1}$,
$~\mu$)  \% backward: error propagation by activation (code) changes
\STATE $\bs{W}_t \gets  \mbox{\bf updateWeights}(   ~\bs{W}_{t-1},~\bs{x}_t,~y_t,~\bs{C},~\mu,~\eta,~Mem$) 
\ENDWHILE
 \STATE \textbf{return} $\bs{W}_t$
 }
 \end{algorithmic}
 \label{alg:AM1}
\end{algorithm}
\vspace{-0.1in}
\end{small}

\begin{small}
\begin{algorithm}[ht]
\caption{  Activation Propagation (Code Update) Steps
}
\label{alg:AM2}
{\bf  encodeInput}($~\bs{x}$,$~\bs{W}$)
 \begin{algorithmic}[1] 
{\small 
\STATE $\bs{c}^0 = \bs{x}$
\FOR{$l=1$ to $L$}   
 \vspace{0.02in}
 \STATE $ \bs{c}^l =  \bs{W}^l \sigma_{l-1}(\bs{c}^{l-1})$\\ 
 \% $\sigma_0(\bs{x})=\bs{x}, \sigma_l(\bs{x})=ReLU(\bs{x}) ~for~ l=1,...,L$
   \vspace{0.01in}
 \ENDFOR 
  \STATE \textbf{return}  $\bs{C}$
  }
 \end{algorithmic}
 \vspace{0.05in} 
 {\bf updateCodes}($~\bs{C}$,$~\bs{y}$,$~\bs{W}$,$~\lambda_C$,$~\mu$)
 \begin{algorithmic}[1]
{\small
 \STATE $\bs{c}^L \leftarrow \text{Solve Problem}$ (\ref{eq:c_L_update}),   $~~\bs{c}^0 = \bs{x}$
 \FOR{$l=L-1$ to $1$}  
   \vspace{0.01in}
 \STATE $\bs{c}^l \leftarrow \text{Solve Problem}$ (\ref{eq:c_l_update})
 \vspace{0.01in}
\ENDFOR 
 \STATE \textbf{return} $\bs{C}$
 }
\end{algorithmic}
\end{algorithm}
\end{small}

\begin{small}
\begin{algorithm}[ht]
\caption{ Weight  and Memory Update Steps
}
\label{alg:AM3}
 {\bf updateWeights}$(~\bs{W},~\bs{x},~y,~\bs{C},~\mu,~\eta,~Mem)$\\
 \begin{algorithmic}[1]
 {\small
 \STATE $\bs{W}^{L+1} = \bs{W^{L+1}} - \eta\nabla_{\bs{W}}{\cal{L}}(y, \sigma_L(\bs{c}^L), \bs{W^{L+1}})$
\FOR{$l=1$ to $L$} 
\IF{$Mem$}
\STATE $(\bs{A^l}_t,\bs{B^l}_t) \gets$ $ \mbox{\bf updateMemory}(~\bs{A^l}_{t-1}$, $~\bs{B^l}_{t-1}$,$~\bs{C^l}$)
   \STATE \%
$  \hat{f}^l(\bs{W^l}) \equiv
Tr(\boldsymbol{W}^T\boldsymbol{W}\boldsymbol{A}^l) - 2 Tr(\boldsymbol{W}^T\boldsymbol{B}^l)$ 
   \STATE
        $\bs{W}^{l} = \arg\min_W \hat{f}^l(\bs{W})$
   \ELSE
        \STATE \%(parallel) local update of each layer  weights,
        \STATE \%({\em independently of other layers (unlike backprop)}
         \STATE $\bs{W}^l \gets $ $ \mbox{\bf SGD}(~\bs{W^l},~\bs{x},~y,~\bs{C^l},~\mu,~\eta)$
\ENDIF
\ENDFOR
\STATE \textbf{return}  $\bs{W} $
}
\end{algorithmic}
  \vspace{0.05in}
  {\bf updateMemory}($~\bs{A}$, $~\bs{B}$,$~\bs{C}$)
  \begin{algorithmic}[1]
{\small  
\FOR{$l=1$ to $L$}  
 \STATE 
$ \bs{a} = \sigma_{l-1}(\bs{c}^{l-1}) $, $~\bs{A}^{l} \gets \bs{A}^{l} + \bs{a}\bs{a}^T$, $~\bs{B}^{l} \gets \bs{B}^{l} + \bs{c}^l\bs{a}^T$
\ENDFOR
 \STATE \textbf{return}  $\bs{A}, \bs{B}$
 }
\label{as:memory_update}
\end{algorithmic}
\end{algorithm}
\vspace{-0.1in}
\end{small}

\noindent{\bf Activation propagation: forward and backward passes.} In an online setting,  
we only have access to the current sample $\bs{x}_t$ at time $t$, and thus can only compute  the corresponding codes $\bs{c}_t^l$ using the weights computed so far.
Namely, given input $\bs{x}_t$, we  compute the last-layer activations $\bs{a}_t^L=\sigma_L(\bs{c}^L_t)$ in a forward pass,  propagating activations from input to the last layer, and make  a prediction about $y_t$, incurring the loss ${\cal{L}}(y_t,\bs{a}_t^L,\bs{W}^{L+1})$. We now  propagate this error back to all activations.   This is achieved by solving a sequence of optimization problems:  
\begin{equation}
\begin{array}{ll}
 \bs{c}^L = & \arg \min_{\bs{c}}      {\cal{L}}(y,\sigma_L(\bs{c}),\bs{W}^{L+1})\\
& 
\quad+ \quad\mu ||\bs{c}  - \bs{W}^L \sigma_{L-1}(\bs{c}^{L-1})||_2^2
\end{array}
\label{eq:c_L_update}
\end{equation}
 \vspace{-0.05in}
\begin{equation}
\begin{array}{ll}
\bs{c}^l = &\arg \min_{\bs{c}}        \mu ||\bs{c}^{l+1} - \bs{W}^{l+1} \sigma_l(\bs{c})||^2_2 \\
 &\quad+\quad\mu ||\bs{c} - \bs{W}^l \sigma_{l-1}(\bs{c}^{l-1})||_2^2, 
\end{array}
\label{eq:c_l_update}
\end{equation}
 \vspace{-0.05in}
for $l=L-1,...,1$. 

\noindent{\bf Weights Update Step.} Different online (stochastic) optimization methods can be applied to update the weights at each layer, using a 
 {\em surrogate}  objective function defined more generally than in  \cite{mairal2009online} as follows:  $\hat{f}_{[t':t]}(\bs{W}) = f(\bs{W},\bs{C}_{[t':t]} )$, where $f$ is defined in eq. \ref{obj1} and $\bs{C}_{[t':t]}$ denotes  codes for all samples from time $t'$ to   time $t$,  computed at previous iterations. 
 When $t'=1$, we  simplify the notation to $\hat{f}_{t}(\bs{W})$, and when $t'=t$, the surrogate is the same as  the true objective  on the current-time  codes $f(\bs{W},\bs{C}_t )$.
The surrogate objective decomposes into $L+1$ independent  terms,  $\hat{f}_t(\bs{W}) = \sum_{l=1}^{L+1} \hat{f}_t^l(\bs{W}^l)$,
which allows for {\em  parallel weight  optimization  across all layers}: 
\vspace{-.2cm}
\begin{equation*}
\bs{W}^{L+1} =  \arg\min_{\bs{W}}\left\{\hat{f}^{L+1}_t(\bs{W})\equiv\sum_{i=1}^t {\cal{L}}(y_i, \sigma_L(\bs{c}_i^L), \bs{W})\right\}.
\end{equation*}
For layers $l=1,...,L$, we have  
\vspace{-.2cm}
\begin{equation}
\bs{W}^l  = \arg\min_{\bs{W}}\left\{\hat{f}^l_t(\bs{W})\equiv \mu  \sum_{i=1}^t \|\bs{c}_i^l - \bs{W}\sigma_{l-1}(\bs{c}_{i}^{l-1})\|^2_2\right\}.\label{eq:weight_update}
\end{equation}
In general, computing a surrogate  function with $t' < t$ would require storing all samples and codes in that time interval.  Thus, for the  $\bs{W}^{L+1}$ update, we always  use   $t'=t$ (current sample), and optimize  ${f}^{L+1}(\bs{W})$     via stochastic gradient descent (SGD) 
(step 1 in {\bf updateWeights},  Algorithm \ref{alg:AM3}).
However,  in    case of quadratic loss (intermediate layers), we have more options. One is to use SGD again, or its adaptive-rate version such as Adam. This option is selected when $Mem=False$ is passed to {\bf updateWeights} function in Algorithm \ref{alg:AM3}.
We call that method {\em AM-Adam}.

Alternatively, we can use the memory-efficient surrogate-function computation as in \cite{mairal2009online}, where $t'=1$, i.e. the surrogate function accumulates the memory of all previous samples and codes, as described below; we hypothesize that such an approach, here called {\em AM-mem}, can be useful in continual learning as a potential mechanism to alleviate the catastrophic forgetting issue.

\noindent{\bf Co-Activation Memory}. We now summarize the memory-based approach. Denoting activation in layer $l$ as $\bs{a}^l=\sigma_l(\bs{c}^l)$, and following \cite{mairal2009online}, we can rewrite the above objective in eq. \ref{eq:weight_update} using the following: 
\vspace{-0.1in}
\begin{eqnarray}
\sum_{i=1}^t   ||
\boldsymbol{c}_i^l - \boldsymbol{W} \boldsymbol{a}_i^l
||_2^2  =
Tr(\boldsymbol{W}^T\boldsymbol{W}\boldsymbol{A}_t^l) - 2Tr(\boldsymbol{W}^T\boldsymbol{B}_t^l),
\label{eqn:Wobj}
\end{eqnarray}
  where $\bs{A}^l_t = \sum_{i=1}^t \boldsymbol{a}_i^{l-1} (\boldsymbol{a}_i^{l-1})^T$ and $\bs{B}^l_t = \sum_{i=1}^t \boldsymbol{c}_i^l (\boldsymbol{a}_i^{l-1} )^T$ are the ``memory''
  matrices (i.e. {\em co-activation memories}), compactly representing the accumulated strength of co-activations in each layer (matrices $\bs{A^l_t}$, i.e. covariances) and across consecutive layers (matrices $\bs{B^l_t}$, or cross-covariances).  
  At each iteration $t$, once the new input sample $\boldsymbol{x}_t $ is encoded,
    the matrices are updated ({\bf updateMemory} function, Algorithm \ref{alg:AM3}) as
    $$\boldsymbol{A_t} \gets \boldsymbol{A_t} + \boldsymbol{a}_t^{l-1} (\boldsymbol{a}_t^{l-1})^T ~\text{and} ~\boldsymbol{B}
\gets
\boldsymbol{B_t} + \boldsymbol{c}_t^l (\boldsymbol{a}_t^{l-1} )^T.$$ 
It is important to note that, using memory matrices, we are effectively optimizing the weights at iteration $t$ with respect to all previous samples and their previous linear activations at all layers, without the need for an explicit storage of these examples. Clearly, AM-SGD is even more memory-efficient since it does not require any memory matrices.
Finally, to optimize the quadratic surrogate in eq. \ref{eqn:Wobj}, we follow \cite{mairal2009online} and use \emph{block-coordinate descent}, iterating over the columns of the corresponding weight matrices; however, rather than always iterating until convergence, we make the number of such iterations an additional hyperparameter.

 
 
 \shrink{
 \noindent{\bf Advantages of our online method(s).} Note that,    unlike \cite{taylor2016training}, where both codes and activations are introduced as auxiliary variables, we only use the codes;     unlike  \cite{carreira2014distributed}, besides more general loss function, we use as  auxiliary variables not the (nonlinear) activations but the corresponding linear codes; this is important for defining the   surrogate objectives for online learning, similar to   the online dictionary learning  method of \cite{mairal2009online}. 
 }

\shrink{ {\em previous samples are not stored explicitly}, so that the memory complexity of our approach remains constant w.r.t. the potentially infinite number of samples $n$, and only depends on the maximum dimensionality of the input and hidden layers, $\max_l m_l$.}

\vspace{-0.1in}
\section{Theoretical analysis}
\vspace{-0.05in}
\label{sec:theory}
We will next provide theoretical convergence analysis for a general stochastic alternating minimization (AM) scheme.  Under certain assumptions that we will discuss, the algorithms proposed in the previous section fall into the category of approaches that comply with these guarantees, although our theory is applicable to a wider family of AM algorithms. {\em To the best of our knowledge, we provide the first theoretical convergence guarantees of AM in the stochastic setting}.


\noindent{\bf Setting.} Let in general $f(\bs\theta_1,\bs\theta_2,\dots,\bs\theta_K)$ denote the function to be optimized using AM, where in the $i^{\text{th}}$ step of the algorithm, we optimize $f$ with respect to $\bs\theta_i$ and keep other arguments fixed. Let $K$ denote total number of arguments. For the theoretical analysis, we consider a smooth approximation to $f$ as done in the literature~\cite{10.1007/978-3-540-74958-5_28,DBLP:conf/esann/LangeZHV14}.

%
%
Let $\{\bs \theta_1^{*},\bs \theta_2^{*}, \dots,\bs\theta_K^{*}\}$ denote the global optimum of $f$ computed on the entire data population. For the sake of the theoretical analysis we assume that the algorithm knows the lower-bound on the radii of convergence $r_1,r_2,\dots,r_K$ for $\bs \theta_1, \bs \theta_2,\dots, \bs \theta_K$.\footnote{This assumption is potentially easy to eliminate with a more careful choice of the step size in the first iterations.} Let $\nabla_if^1$ denote the gradient of $f$ computed for a single data sample $(\bs x, \bs y)$ and taken with respect to the $i^{\text{th}}$ argument of the function $f$ (weights or codes from Algorithm~\ref{alg:AM}). In the next section, we refer to $\nabla_i f(\cdot)$ as the gradient of $f$ with respect to ${\bs\theta_i}$ computed for the entire data population, i.e. an infinite number of samples (``oracle gradient''). We assume in the $i^{\text{th}}$ step ($i = 1,2,\dots,K$), the AM algorithm performs the update:
\vspace{-0.05in}
\begin{equation}
\bs \theta_i^{t+1} \!=\! \Pi_i(\bs \theta_i^t - \eta^{\tau}\nabla_if^1(\bs\theta_1^{t+1},\dots,\bs\theta_{i-1}^{t+1},\bs\theta_{i}^{t}, \bs\theta_{i+1}^{t},\dots,\bs\theta_K^t)),
      \label{eq:update}
      \end{equation}
where $t$ denotes time, $\Pi_i$ denotes the projection onto the Euclidean ball $B_2(\frac{r_i}{2},\bs \theta_i^0)$ of some given radius $\frac{r_i}{2}$ centered at the initial iterate $\bs \theta_i^0$. Thus, given any initial vector $\bs \theta_i^{0}$ in the ball of radius $\frac{r_i}{2}$ centered at $\bs \theta_i^{*}$, we are guaranteed that all iterates remain within an $r_i$-ball of $\bs \theta_i^{*}$. This is true for all $i = 1,2,\dots,K$. The re-projection step of eq.~\ref{eq:update} implies that
starting close enough to the optimum and taking small steps leads to convergence rate of Theorem~\ref{thm:erroruN}. The radiuses dictate how convergence is affected if the iterates stray further from the
optimum through the variable $\sigma^2$ defined before that theorem. 

\begin{remark}
The difference between the AM scheme we analyze and the   Algorithm~\ref{alg:AM} can be summarized as follows:  i) only a single SGD step is taken
with respect to weights and then codes (while Algorithm~\ref{alg:AM} can optimize codes till convergence at each
iteration); ii) gradient direction is approximated with respect to a single data sample (in practice,
Algorithm~\ref{alg:AM} uses mini-batches), and iii) re-projection step is included, unlike in Algorithm~\ref{alg:AM}.

We argue that the general AM scheme analyzed here  
leads to \textit{the worst-case theoretical guarantees} with respect to the original setting from 
Algorithm \ref{alg:AM}, i.e. we expect the convergence rate for the original setting to be no worse than the one dictated by the obtained guarantees. This is because we allow only a single stochastic update (i.e. computed on a single data point) with respect to an appropriate argument (when keeping other arguments fixed) in each step of AM, whereas in Algorithm \ref{alg:AM} and related schemes in the literature, one may increase the size of the data mini-batch in each AM step (semi-stochastic setting). The convergence rate in the latter case is typically better~\cite{Nesterov:2014:ILC:2670022}. Finally, note that the analysis does not consider running the optimizer more than once before changing the argument of an update, e.g., when obtaining sparse code $\bs c$ for a given data point $(\bs x, \bs y)$ and fixed weights. We expect this to have a minor influence on the convergence rate as our analysis specifically considers a local convergence regime, where we expect that running the optimizer once produces good enough parameter approximations. Moreover, note that by preventing each AM step to be performed multiple times, we analyze a more stochastic (noisier) version of parameter updates.
\end{remark}

\noindent{\bf Statistical guarantees for AM algorithms.}
The theoretical analysis we provide here is an extension to the AM setting of recent work on statistical guarantees for the EM algorithm~\cite{balakrishnan2017}.

We first discuss necessary assumptions that we make. Let $L (\bs \theta_1,\bs \theta_2, \dots,\bs\theta_K) = -f(\bs \theta_1,\bs \theta_2, \dots,\bs\theta_K)$ and denote $L^*_d(\bs\theta_d)=L(\bs \theta_1^{*},\bs\theta_2^{*},\dots,\bs\theta_{d-1}^{*},\bs\theta_d,\bs\theta_{d+1}^{*},\dots,\bs\theta_{K-1}^{*},\bs\theta_K^{*})$. Let $\Omega_1,\Omega_2,\dots,\Omega_K$ denote non-empty compact convex sets such that for any $i = \{1,2,\dots,K\}, \bs\theta_i\in\Omega_i$. The following three assumptions are made on $L^*_d(\bs\theta_d)$ ($d = 1,2,\dots,K$) and the objective function $L (\bs \theta_1,\bs \theta_2, \dots,\bs\theta_K)$. 


\begin{assumption}[{\em Strong concavity}]
The function $L^*_d(\bs\theta_d)$ is strongly concave for all pairs $(\bs \theta_{d,1},\bs \theta_{d,2})$ in the neighborhood of $\bs \theta_d^{*}$. That is
\begin{eqnarray*}
L^*_d(\bs\theta_{d,1}) - L^*_d(\bs\theta_{d,2}) - \left<\nabla_d L^*_d(\bs\theta_{d,2}), \bs\theta_{d,1}-\bs\theta_{d,2}\right>\\
\leq -\frac{\lambda_d}{2}\|\bs\theta_{d,1} - \bs\theta_{d,2}\|^2_2,
\end{eqnarray*}
where $\lambda_d > 0$ is the strong concavity modulus. 
\label{def:strongconN}
\end{assumption}

\begin{assumption}[{\em Smoothness}]
The function $L^*_d(\bs\theta_d)$ is $\mu_d$-smooth for all pairs $(\bs\theta_{d,1},\bs\theta_{d,2})$. That is
\begin{eqnarray*}
L^*_d(\bs\theta_{d,1}) - L^*_d(\bs\theta_{d,2}) - \left<\nabla_d L^*_d(\bs\theta_{d,2}), \bs\theta_{d,1}-\bs\theta_{d,2}\right>\\
\geq -\frac{\mu_d}{2}\|\bs\theta_{d,1} - \bs\theta_{d,2}\|^2_2,
\end{eqnarray*}
where $\mu_d > 0$ is the smoothness constant. 
\label{def:smoothN}
\end{assumption}

Next, we introduce the gradient stability (GS) condition that holds for any $d$ from $1$ to $k$. 

\begin{assumption}[{\em Gradient stability (GS)}]
We assume $L(\bs\theta_1,\bs\theta_2\dots,\bs\theta_K)$ satisfies GS ($\gamma_d$) condition, where $\gamma_d \geq 0$, over Euclidean balls $\bs\theta_1 \in B_2(r_1,\bs\theta_1^{*}), \dots,\bs\theta_{d-1}\in B_2(r_{d-1},\bs\theta_{d-1}^{*}),\bs\theta_{d+1}\in B_2(r_{d+1},\bs\theta_{d+1}^{*}),\dots, \bs\theta_{K} \in B_2(r_{K},\theta_{K}^{*})$ of the form
\vspace{-0.07in}
\[ 
\|\nabla_d L^*_d(\bs\theta_d) -\nabla_d L(\bs\theta_1, \bs\theta_2,\dots,\bs\theta_K)\|_2\leq \gamma_d\sum_{\substack{i=1 \\ i\neq d}}^K\|\bs\theta_i - \bs\theta_i^{*}\|_2.
\]
\vspace{-0.07in}
\label{def:GSN}
\vspace{-0.2in}
\end{assumption}

We also define the following bound $\sigma$ on the expected value of the norm of the gradients of our objective function (commonly done in the stochastic gradient descent convergence theorems as well). Define $\sigma = \sqrt{\sum_{d=1}^K\sigma_{d}^2}$ where 
\begin{eqnarray*}
\sigma_{d}^2 = \sup\{&&\hspace{-0.27in}\mathbb{E}[\|\nabla_d L_1(\bs\theta_1,\bs\theta_2,\dots,\bs\theta_K)\|_2^2] : \\&&\bs\theta_1 \in B_2(r_1,\bs\theta_1^{*}) \ldots \bs\theta_K \in B_2(r_k,\bs\theta_k^{*}) \}
\end{eqnarray*}

The following theorem then gives a recursion on the expected error obtained at each iteration of Algorithm 1.
\begin{theorem}
Given the stochastic AM gradient iterates of the version of Algorithm~\ref{alg:AM} given in eq.~\ref{eq:update} with decaying step size $\{\eta^t\}_{t=0}^{\infty}$ and $\gamma < \frac{2\xi}{3(K-1)}$, the error at iteration $t+1$ satisfies recursion
\vspace{-0.1in}
\begin{eqnarray}
\mathbb{E}\left[\sum_{d=1}^K\|\bs\Delta^{t+1}_d\|_2^2\right] &\leq& (1-q^t)\mathbb{E}\left[\sum_{d=1}^K\|\bs\Delta^{t}_d\|_2^2\right] \nonumber \\&&+ \frac{(\eta^t)^2}{1-(K-1)\eta^t\gamma}\sigma^2,
\end{eqnarray}
where $\bs\Delta^{t+1}_d \coloneqq \bs\theta_d^{t+1} - \bs\theta_d^{*}$ for $d = 1,2,\dots,K$, $\gamma \coloneqq \max_{i=1,2,\dots,K}\gamma_i$, $q^t = 1 - \frac{1-2\eta^t\xi+2\eta^t\gamma(K-1)}{1-(K-1)\eta^t\gamma}$, and $\xi \coloneqq \min_{i = 1,2,\dots,K}\frac{2\mu_i\lambda_i}{\mu_i + \lambda_i}$.
\label{thm:erroruN}
\end{theorem}
The recursion in Theorem \ref{thm:erroruN} is expanded in the Supplementary Material to prove the final convergence theorem stated as follows:  

\begin{theorem}
Given the stochastic AM gradient iterates of the version of Algorithm~\ref{alg:AM} given in eq.~\ref{eq:update} with decaying step size $\eta^t = \frac{3/2}{[2\xi-3\gamma(K-1)](t+2) + \frac{3}{2}(K-1)\gamma}$ and assuming that $\gamma < \frac{2\xi}{3(K-1)}$, the error at iteration $t+1$ satisfies
\vspace{-0.1in}
\begin{eqnarray}
\!\!\!\!\!\!\!\!\!\mathbb{E}\left[\sum_{d=1}^K\|\bs\Delta^{t+1}_d\|_2^2\right] &\leq& \mathbb{E}\left[\sum_{d=1}^K\|\bs\Delta^0_d\|_2^2\right]\left(\frac{2}{t+3}\right)^{\frac{3}{2}}\nonumber\\&& \!\!\!\!\!\!\!\!\!\!+ \sigma^2\frac{9}{[2\xi-3\gamma(K-1)]^2(t+3)},
\end{eqnarray}
where $\bs\Delta^{t+1}_d \coloneqq \bs\theta_d^{t+1} - \bs\theta_d^{*}$ for $d = 1,2,\dots,K$, $\gamma \coloneqq \max_{i=1,2,\dots,K}\gamma_i$, and $\xi \coloneqq \min_{i = 1,2,\dots,K}\frac{2\mu_i\lambda_i}{\mu_i + \lambda_i}$.
\label{lem:errorfinal}
\end{theorem}

\begin{small}
\begin{figure*}[t]
\begin{multicols}{2}
    \hspace{-0.2in}
    \includegraphics[width=3.6in,height=2.9in]{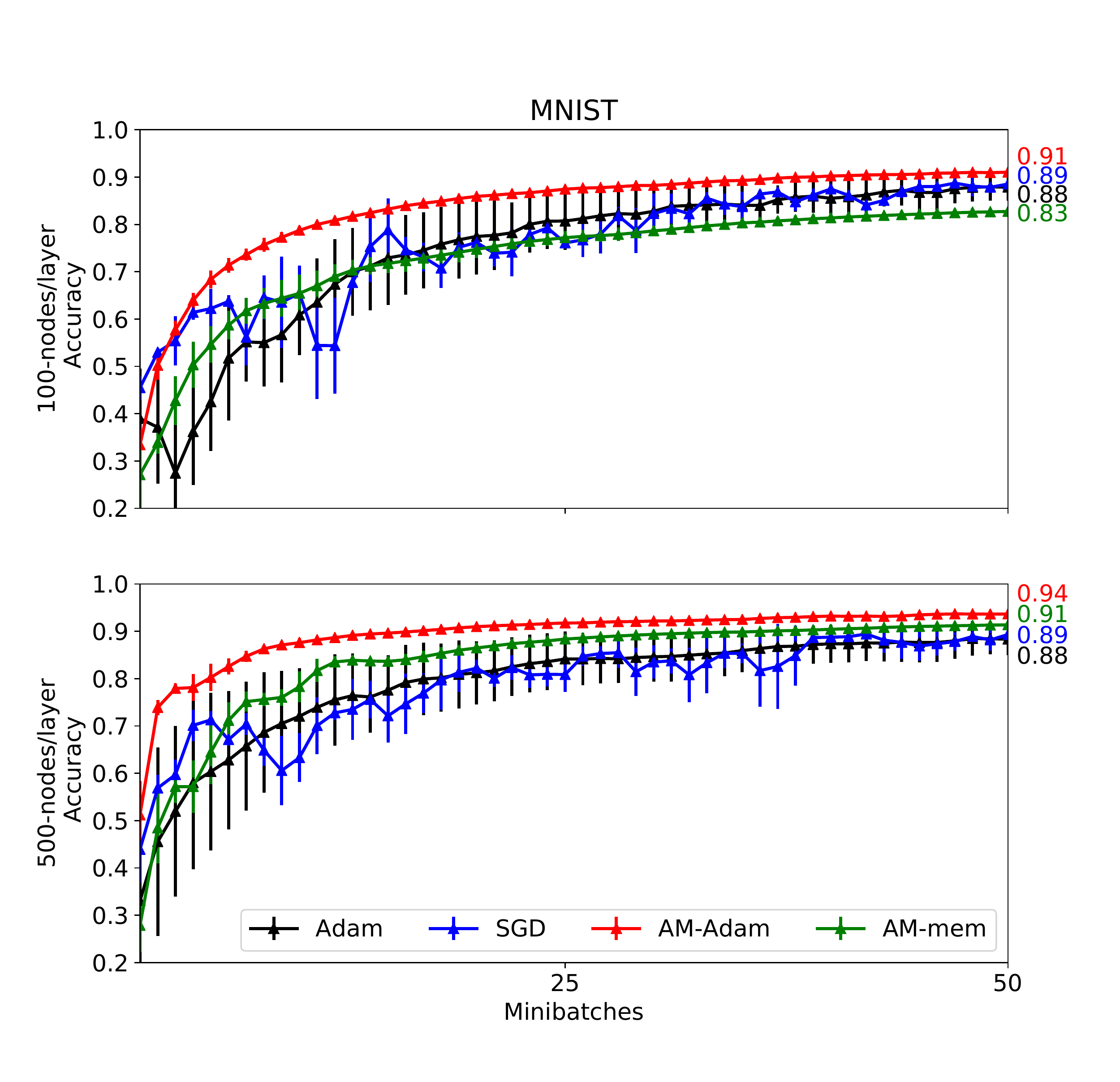}
    \par \vspace{-0.32in} \caption{\small    MNIST (fully-connected nets, 2 layers): online methods,  first epoch;  50 mini-batches, 200 samples each.}
    \label{fig:MNIST1}
  \hspace{-0.26in}
      \includegraphics[width=3.6in,height=2.9in]{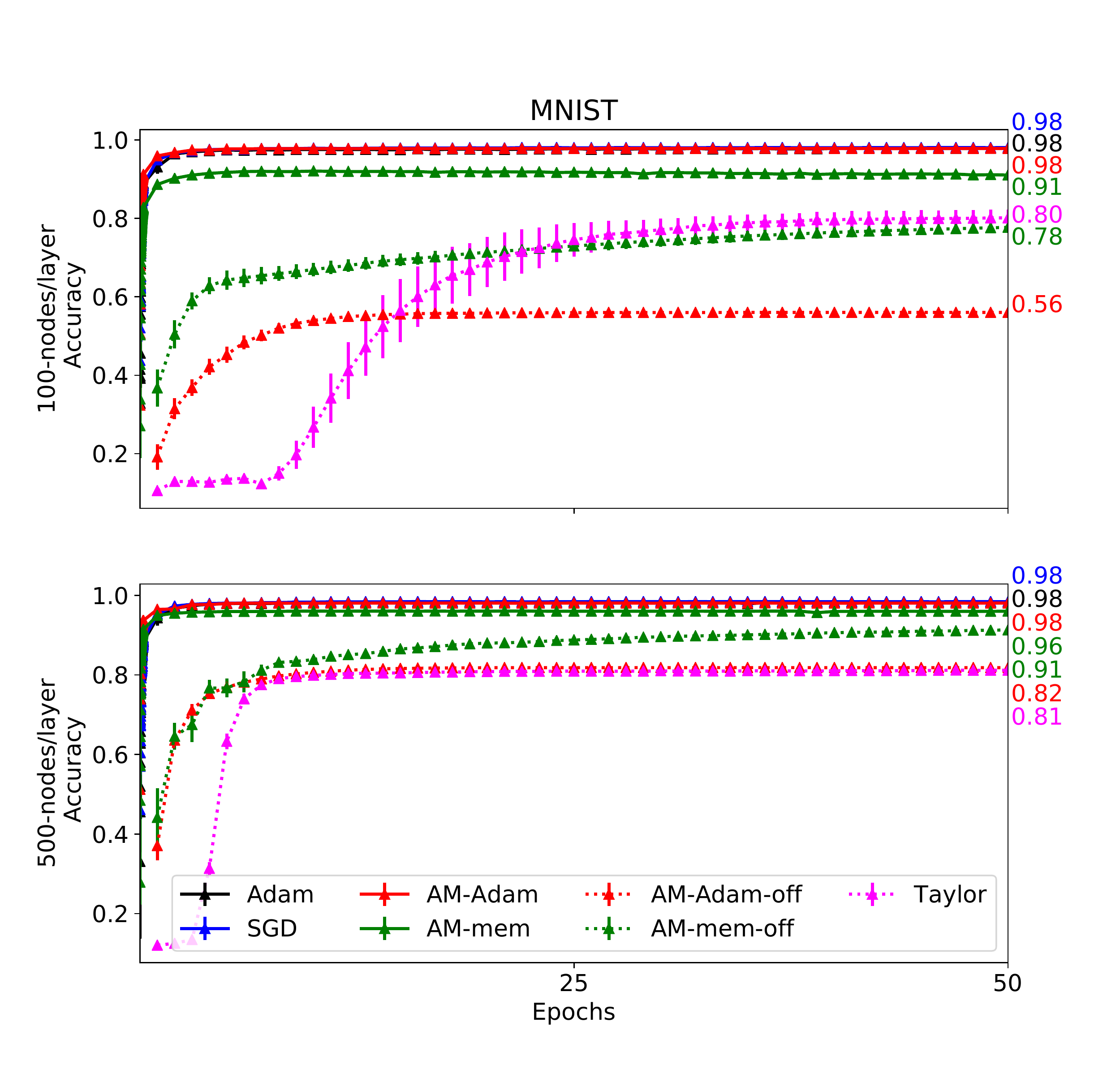}
      \par \vspace{-0.32in} \caption{\small MNIST (fully-connected nets, 2 layers):  online vs. offline methods vs. Taylor's ADMM, 50 epochs.     }
      \label{fig:MNIST2}
          \hspace{-0.21in}              
\end{multicols}
\vspace{-0.4in}
\end{figure*}
\end{small} 

\begin{small}
\begin{figure*}[t]
\begin{multicols}{2}
    \hspace{-0.2in}
    \includegraphics[width=3.6in,height=2.9in]{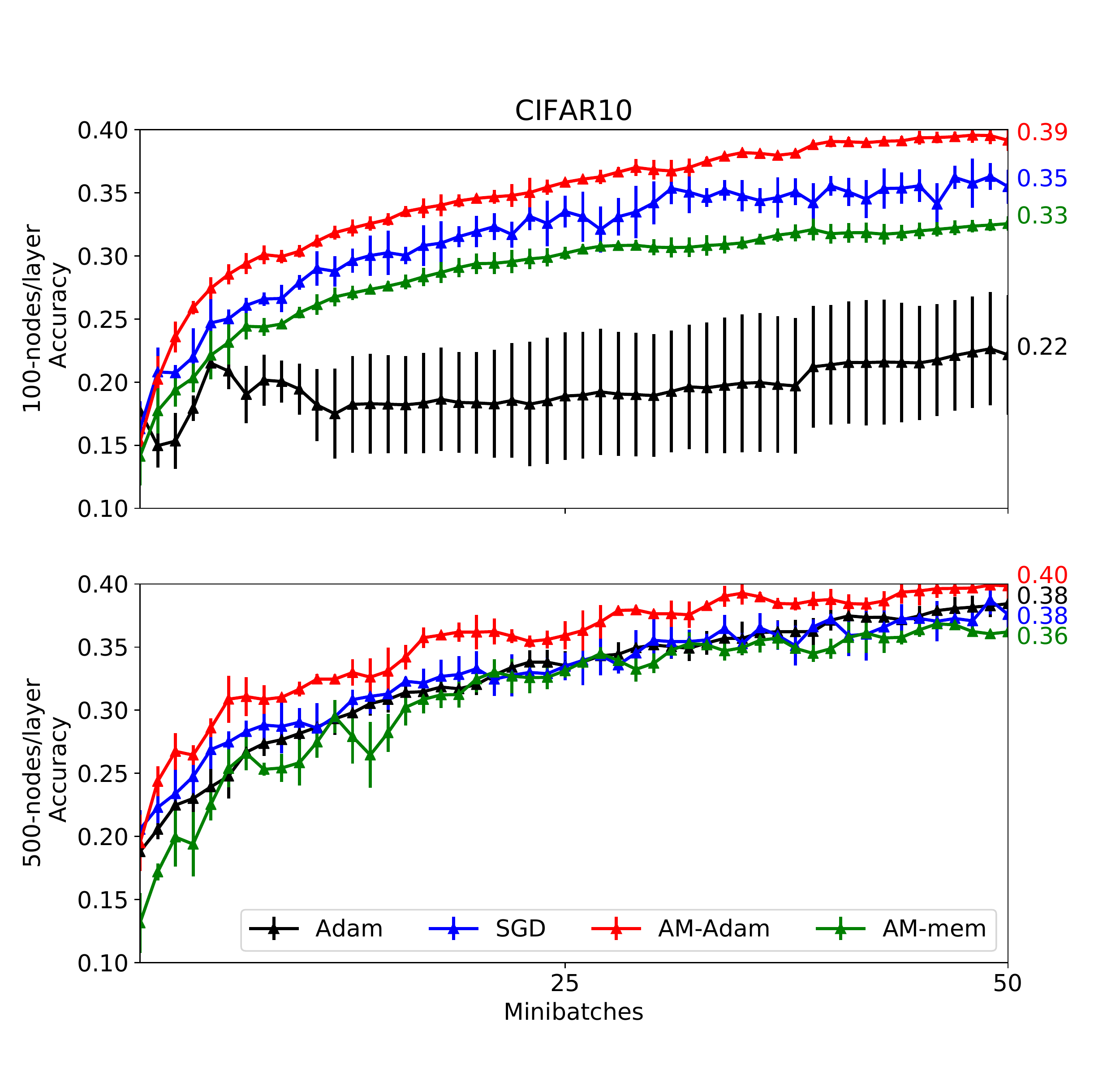}\par \vspace{-0.32in} \caption{\small  CIFAR10 (fully-connected nets): online methods, 1st epoch.  2 hidden layers with 100 (top) and 500 (bottom)   units each;    250 mini-batches, 200 samples each.}\label{fig:CIFAR1}
  \hspace{-0.26in}
      \includegraphics[width=3.6in,height=2.9in]{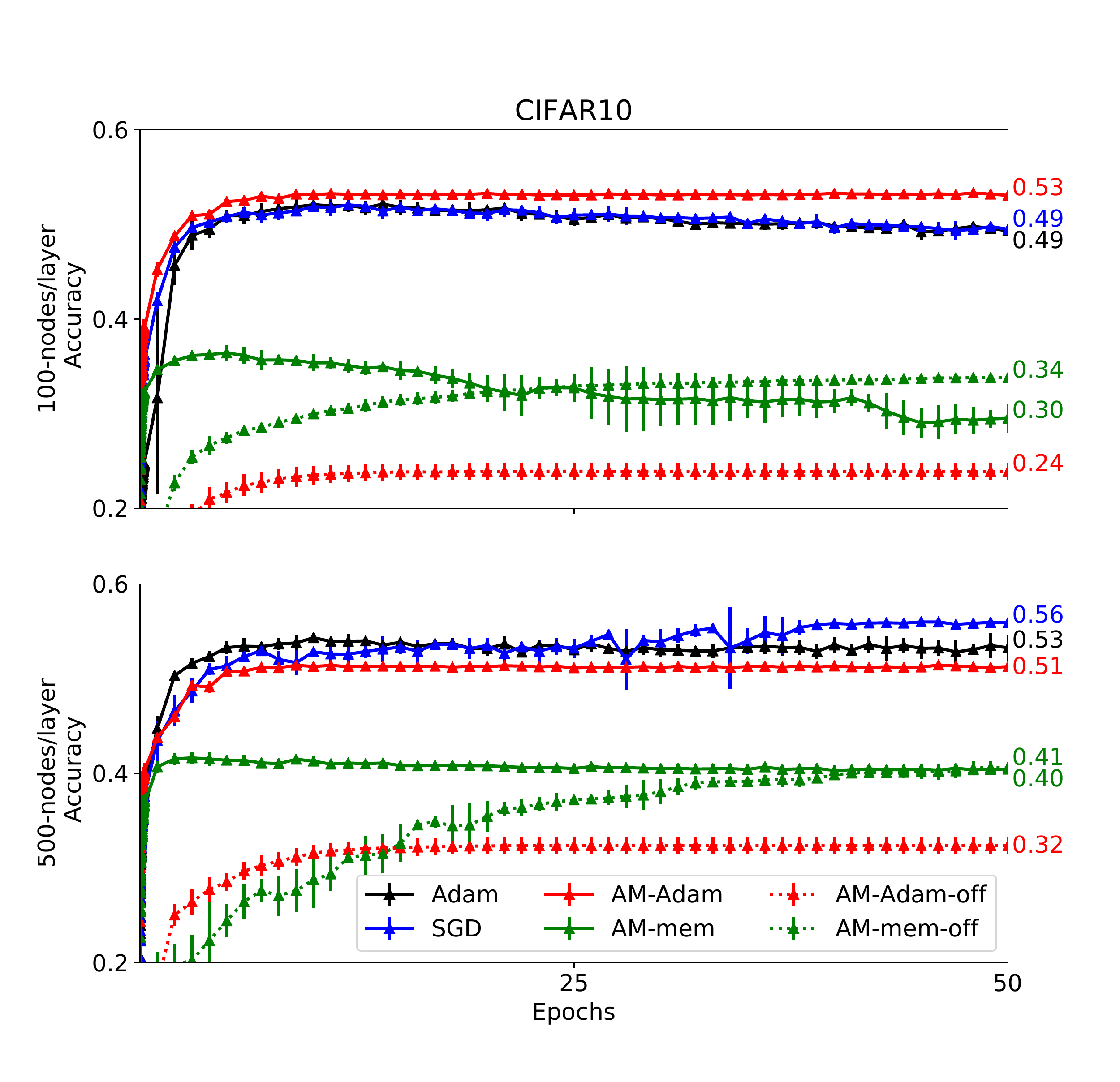}\par \vspace{-0.32in} \caption{ CIFAR10 (fully-connected networks): online vs. offline, 50 epochs. Similar experiments to Figure 2.}\label{fig:CIFAR2}
   \hspace{-0.21in}
   \end{multicols}
 \vspace{-0.3in}
\end{figure*}
\end{small}

\shrink{
}

\vspace{-0.15in}
\section{Experiments}
\vspace{-0.05in}
\shrink{
NOTE: A possible intuition behind AM learning faster early - CP note on "shortcut in (W,C) space.
 }

We compare on several datasets (MNIST, CIFAR10, HIGGS) our online alternating minimization algorithms,   {\em  AM-mem} and {\em  AM-Adam} (using mini-batches instead of single samples at each time point), against  backrop-based online methods, SGD and   Adam \cite{kingma2014adam}, as well as against the offline auxiliary-variable ADMM method of \cite{taylor2016training}, using code provided by the authors\footnote{We choose Taylor's ADMM among several auxiliary methods proposed recently, since it was the only one capable of handling very large datasets due to massive data parallelization; also, some other methods were  not designed for classification task, e.g. \cite{carreira2014distributed} trained autoencoders, \cite{zhang2016efficient} learned hashing. }, and against the two offline versions of our methods, {\em AM-Adam-off} and {\em AM-mem-off}, which simply treat the training dataset as a single  minibatch, i.e. one AM iteration is equivalent to one epoch over the training dataset. 
 All our  algorithms were implemented in PyTorch \cite{paszke2017automatic}; we also used PyTorch implementation of {\em SGD} and {\em Adam}. Hyperparameters used for each method 
 were optimized by  grid search on  a validation subset of training data. Most results were averaged over at least 5 different weight initializations.
 
Note that most of the prior auxiliary-variable methods were evaluated only on fully-connected networks  \citep{carreira2014distributed,taylor2016training,zhang2016efficient,zhang2017convergent,zeng2018block,Askari2018}, while  {\em we also experiment with RNNs and CNNs, as well as with  discrete (nondifferentiable) networks}. 


\noindent{\bf Fully-connected nets: MNIST, CIFAR10, HIGGS.} 
We experiment with fully-connected networks on the standard MNIST \cite{lecun1998mnist} dataset, consisting of $28\times 28$ gray-scale images of hand-drawn digits, with 50K samples, and a test set of 10K samples.   We evaluate two different 2-hidden-layer architectures, with equal hidden layer sizes of 100 and 500, and ReLU activations. Figure \ref{fig:MNIST1}  zooms in on the performance of the online methods,  {\em AM-Adam}, {\em AM-mem}, {\em SGD}  and {\em Adam}, over 50 minibatches of size 200 each. We observe that, on both architectures,   
{\em AM-Adam is comparable to (in early stages, even slightly better than)   SGD and Adam}, while {\em AM-mem} is comparable with them on the larger architecture, and falls between {\em SGD} and {\em Adam} on the smaller one. Next,  Figure \ref{fig:MNIST2} continues  to 50 epochs, now including the offline methods (which require at least 1 epoch over the full dataset, by definition). Our {\em AM-Adam} method matches  {\em SGD}  and {\em Adam}, reaching 0.98 accuracy.  Our second method, {\em AM-mem} only yields 0.91 and 0.96 on the 100-node and 500-node networks, respectively. {\em All offline methods are significantly outperformed by the online ones; e.g., Taylor's ADMM learns very slowly until about 10 epochs, being greatly outperformed even  by our offline versions, but later catches up with offline {\em AM-mem} on the 100-node network; it is still inferior to all other methods on the 500-node architecture.}
 
Figures \ref{fig:CIFAR1} and \ref{fig:CIFAR2} show similar results for the same experiment setting, on the  CIFAR10 dataset 
(5000 training and 10000 test samples). Again, our {\em AM-Adam} performs slightly better than SGD and Adam on the first 50 minibatches (same size 200 as before), and even on 50 epochs for the 1-100 architecture, reaching 0.53 vs 0.49 accuracy of SGD and Adam, but falls a bit behind on the larger 1-500 architecture with 0.51 vs 0.53 and  0.56, respectively. Our second algorithm, {\em AM-mem}, is clearly dominated by all the three methods above.
Also, we ran the two offline AM versions, which were again greatly outperformed by the online methods. {\em In the remaining experiments, we focus on our best-performing method,  online AM-Adam}.

\begin{figure*}[t]
\begin{multicols}{3}
 \hspace{-0.1in}  
 \includegraphics[width=2.3in,height=1.7in]{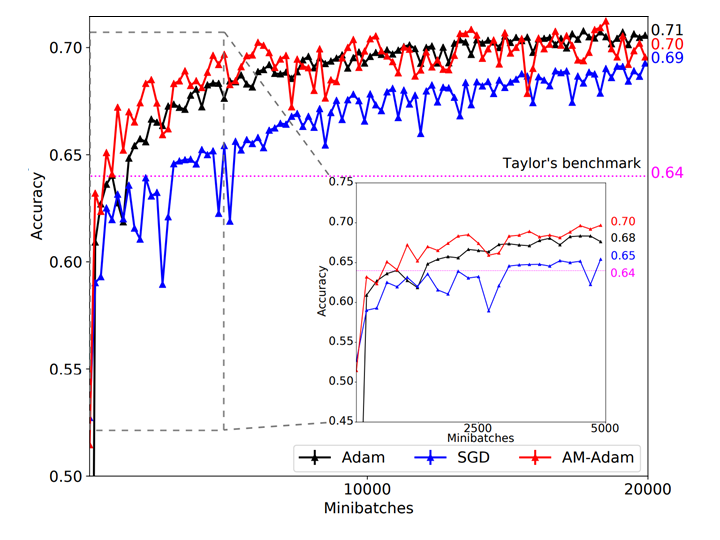} \par \vspace{-0.2in} 
 \caption{\small  HIGGS dataset. 
  }
   \vspace{0.2in} 
 \label{fig:HIGGS1}
 \hspace{-0.2in} 
 \includegraphics[width=2.2in,height=1.7in]{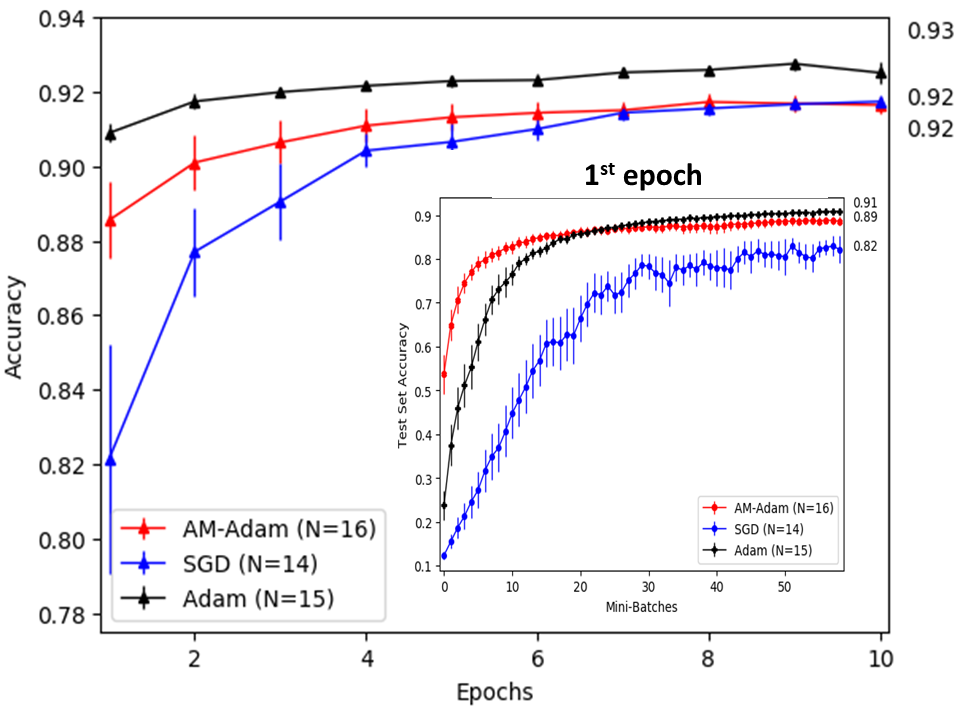} \par \vspace{-0.2in} \caption{\small    RNN-15, Sequential MNIST. }\label{fig:RNN15}
\hspace{-0.3in}              
\includegraphics[width=2.3in,height=1.75in]{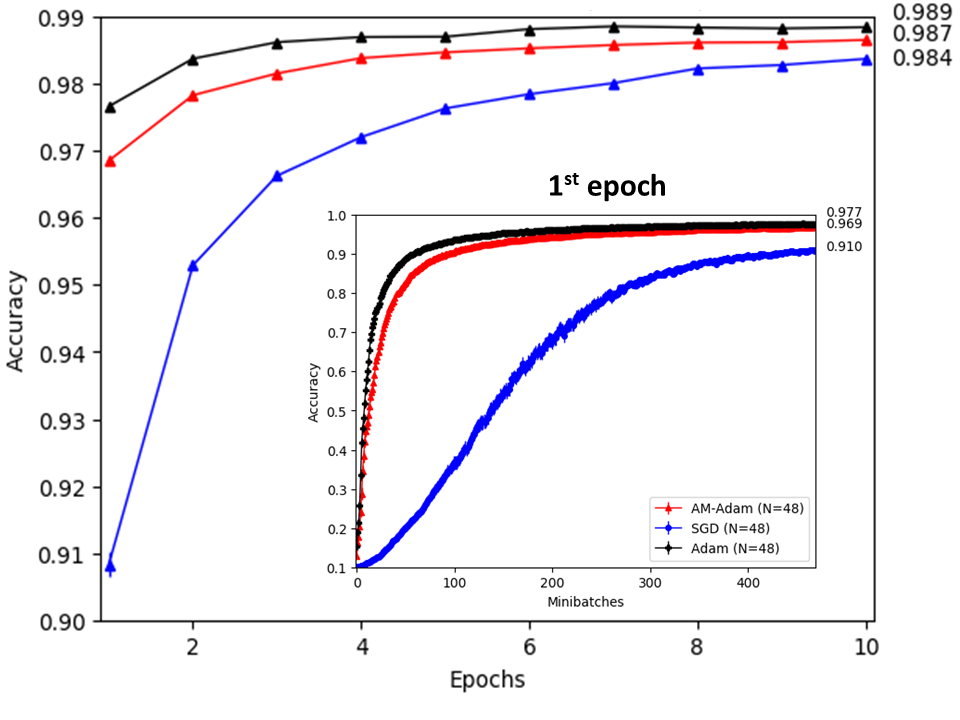} \par \vspace{-0.22in} \caption{\small  CNN: LeNet5, MNIST.}\label{fig:CNN_MNIST}
\end{multicols}
\vspace{-0.3in}
\end{figure*}

\noindent{\bf HIGGS,  fully-connected, 1-300 ReLU network.} In Figure \ref{fig:HIGGS1}, we compare our {\em online AM-Adam} approach against {\em SGD}, {\em Adam} and the offline ADMM method of Taylor,  on a very large  HIGGS dataset, containing   10,500,000 training samples (28 features each) and  500,000 test samples. Each datapoint is labeled as either a signal process producing a Higgs boson or a background process which does not. 
 We use the same architecture (a single-hidden layer network with  ReLU activations and 300 hidden nodes) as in    \cite{taylor2016training}, and the same training/test data sets.
For all online methods, we use  minibatches of size 200, so one epoch over the 10.5M samples equals 52,500 iterations.  

 While Taylor's method was reported to achieve 0.64 accuracy on the whole  dataset (using data parallelization on 7200 cores to handle the whole dataset as a batch) \cite{taylor2016training}, the online methods achieve the same accuracy much faster (less than 1000 iterations/200K samples for our {\em AM-Adam}, and less than 2000 iterations for {\em SGD} and {\em Adam}; within only 20,000 iterations (less than a half of training samples),  {\em AM-Adam}, {\em SGD} and  {\em Adam}   0.70, 0.69 and 0.71, respectively, and continue to improve slowly, reaching  after one epoch,  0.71, 0,71 and 0.72, respectively.     (Our {\em AM-mem} version quickly reached 0.6 together with {\em AM-Adam}, but then slowed down, reaching only 0.61  on the 1st epoch). 
 
In summary,  {\em on HIGGS dataset,  {\em AM-Adam}, SGD and Adam  clearly outperform  Taylor's offline ADMM, while using less than a half of the 1st epoch, and quickly reaching Taylor's 0.64 accuracy benchmark after observing only a tiny fraction (less than 0.01\%) of the 10.5M  dataset. Both Adam and AM-Adam perform very closely, both outperforming   SGD.}

 \noindent{\bf RNN on MNIST.} Next, we evaluate our method on  Sequential MNIST \cite{le2015simple}, where each image is vectorized and fed to the RNN as a sequence of $T=784$ pixels. We use the standard Elman RNN architecture with $tanh$ activations among hidden states and ReLU applied to the output sequence before making a prediction (we use larger minibatches  of 1024 samples to reduce training time).     {\em AM-Adam} was adapted to work on  such RNN architecture (see Appendix for details). 
Figure \ref{fig:RNN15} shows the results using $d=15$ hidden units (see Appendix for $d=50$), averaged over N   weight  initializations, for 10 epochs, with a zoom-in on the first epoch inset.   {\em  AM-Adam performs similarly to Adam in the 1st epoch,  and outperforms SGD up to epoch 6,  matching SGD's performance afterwards.}
 
\noindent{\bf CNN (LeNet-5), MNIST.} Next, we experiment with CNNs, using LeNet-5 \cite{LeCun1998} on  MNIST (Figure \ref{fig:CNN_MNIST}). Similarly to RNN result, AM-Adam clearly outperforms SGD, while being somewhat outperformed by Adam. 

\noindent{\bf Binary nets (nondifferentiable activations), MNIST.}
Finally, to  investigate the ability of our method to handle non-differentiable networks, 
we consider an architecture originally investigated in \cite{lee2015difference} to evaluate another type of auxiliary-variable approach, called Difference Target Propagation (DTP).
The model is a 2-hidden layer fully-connected network (784-500-500-10), whose first hidden layer uses the non-differentiable $sign$ transfer function (while the second hidden layer uses $\tanh$).
Target propagation approaches were motivated by the goal of finding more biologically plausible mechanisms for credit assignment in the brain's neural networks as compared to standard backprop, which, among multiple other biologically-implausible aspects, does not model the neuronal activation propagation explicitly, and does not handle   non-differentiable binary activations (spikes) \cite{lee2015difference,bartunov2018assessing}. 
\begin{wrapfigure}{r}{4.3cm}
\vspace{-0.23in}
\label{fig:nondiff}
\centerline{\includegraphics[width=130pt,height=110pt]{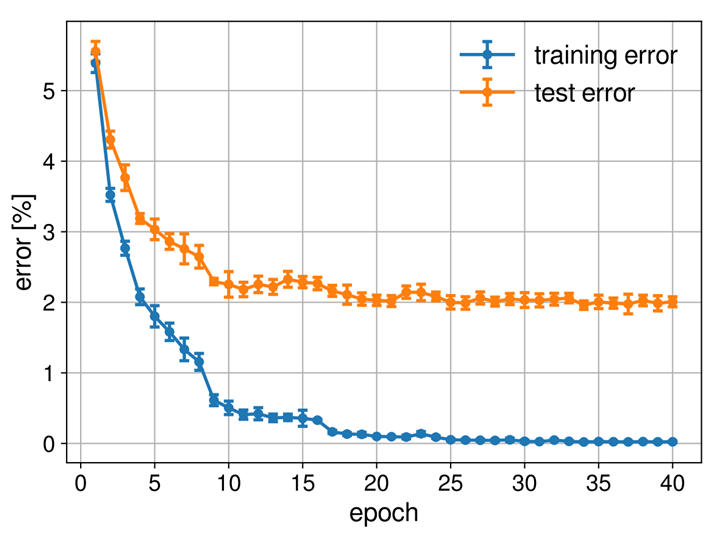} }
\vspace{-0.2in}
\caption{\small Binary net, MNIST.}
\vspace{-0.1in}
\end{wrapfigure}
 In \cite{lee2015difference}, DTP was applied to the above discrete network, and compared to a backprop-based {\em straight-through estimator (STE)}, which simply ignores
the derivative of the step function (which is 0 or infinite) in the back-propagation phase. 
{\em While DTP took about 200 epochs to reach 0.2 error, matching the STE performance (Figure 3 in \cite{lee2015difference}), our AM-Adam with binary activations reaches the same error in less than 20 epochs (Figure 8).} 

\shrink{Finally,  runtime results were mostly comparable between our AM-Adam and Adam/SGD, even without implementing  parallel weight updates which could provide a considerable speed-up as previously demonstrated by \cite{carreira2014distributed} (see Supplementary Material for more details on runtimes).}

\shrink{
Figures \ref{fig:1} and \ref{fig:2} present the results on HIGGS, where we used a single-hidden layer network with ReLU activations and a hidden layer of 300 nodes, as suggested in \cite{Baldi2014}, and also a larger network with 500 nodes. We used this dataset as a simple sanity-check, in order to compare an offline ADMM procedure~\cite{taylor2016training} with our online procedure. As shown in Figure \ref{fig:1}, using the same network architecture, $(AM)^2$ achieves about 63\% accuracy after processing about 300 minibatches of size 100, i.e. after about 30000 samples, which is comparable to 64\% accuracy achieved by ADMM on this very large dataset, comprising 10,500,000 training datapoints of 28 features each, with each datapoint labeled as either a signal process producing a Higgs boson or a background process which does not. These results are for one learning epoch. Figure \ref{fig:2} compares our online procedure with the offline AM procedure, where the whole dataset is treated as one minibatch, and both codes and weights are optimized to convergence using a single global iteration. Interestingly, on this dataset, the online approach seems to outperform its offline counterpart.
}

\vspace{-0.1in}
\section{Conclusions}
\vspace{-0.05in}
We proposed a novel online alternating-minimization approach for neural network training; it  builds upon previously proposed offline methods that break the nested objective into easier-to-solve local subproblems via inserting auxiliary variables corresponding to activations in each layer. Such methods  avoid gradient chain computation and potential issues associated with it, including vanishing gradients, lack of cross-layer parallelization, and difficulties handling non-differentiable nonlinearities.
However, unlike prior art, our approach is online (mini-batch), and thus can handle arbitrarily large datasets and  continual learning settings. We proposed two  variants, AM-mem and AM-Adam, and found that AM-Adam works better.  Also, AM-Adam greatly outperforms offline methods on several datasets and architectures; when compared to state-of-the-art backprop methods such as (standard) SGD and Adam,
 AM-Adam typically  matches  their performance over multiple epochs, and may even  learn somewhat faster initially, in small-data regimes.  AM-Adam also converged faster than another related method, difference target propagation, on a discrete (non-differentiable) network.
Finally, to the best of our knowledge, we are the first to  provide theoretical guarantees for a wide class of online alternating minimization approaches including ours.

\begin{small}
\bibliography{AM}
\bibliographystyle{icml2019}
\end{small}

\appendix

\newpage
\onecolumn

\centerline{\bf \Large Supplementary Material}
\section{Proofs}
Proof of Theorem \ref{lem:errorfinal} relies on Theorem \ref{thm:erroruN}, which in turn relies on Theorem \ref{thm:contractivityG1wideN} and Lemma \ref{lem:contru2N}, both of which are stated below. Proofs of the lemma and theorems follow in the subsequent subsections.

The next result is a standard result from convex optimization (Theorem 2.1.14 in~\cite{Nesterov:2014:ILC:2670022}) and is used in the proof of Theorem~\ref{thm:contractivityG1wideN} below.

Next, we introduce the \textit{population gradient AM operator}, $\mathcal{G}_i(\bs\theta_1,\bs \theta_2,\dots,\bs\theta_K$), where $i = 1,2,\dots,K$, defined as
\vspace{-0.07in}
\[
\mathcal{G}_i(\bs\theta_1,\bs\theta_2,\dots,\bs\theta_K) \coloneqq \bs\theta_i + \eta\nabla_{i} f(\bs\theta_1,\bs\theta_2,\dots,\bs\theta_K),
\]
\vspace{-0.07in}
where $\eta$ is the step size. 

\begin{lemma}
For any $d = 1,2,\dots,K$, the gradient operator $\mathcal{G}_d(\bs\theta_1^{*},\bs\theta_2^{*},\dots,\bs\theta_{d-1}^{*},\bs\theta_{d},\bs\theta_{d+1}^{*},\dots,\bs\theta_{K-1}^{*},\bs\theta_K^{*})$ under Assumption~\ref{def:strongconN} (strong concavity) and Assumption~\ref{def:smoothN} (smoothness) with constant step size choice $0 < \eta \leq \frac{2}{\mu_d + \lambda_d}$ is contractive, i.e.
\begin{equation}
\|\mathcal{G}_d(\bs\theta_1^{*},\dots,\bs\theta_{d-1}^{*},\bs\theta_{d},\bs\theta_{d+1}^{*},\dots,\bs\theta_K^{*}) - \bs\theta_d^{*}\|_2 \leq \left(1 - \frac{2\eta\mu_d\lambda_d}{\mu_d + \lambda_d}\right)\|\bs\theta_d - \bs\theta^{*}_d\|_2
\label{eq:contractivityTN}
\end{equation}
for all $\bs\theta_d \in B_2(r_d,\bs\theta_d^{*})$.
\label{lem:contru2N}
\end{lemma}

The next theorem also holds for any $d$ from $1$ to $K$. Let $r_1, \dots,r_{d-1},r_{d+1},\dots,r_K > 0$ and $\bs\theta_1 \in B_2(r_1,\bs\theta_1^{*}),\dots, \bs\theta_{d-1} \in B_2(r_{d-1},\bs\theta_{d-1}^{*}), \bs\theta_{d+1} \in B_2(r_{d+1},\bs\theta_{d+1}^{*}),\dots,\bs\theta_{K} \in B_2(r_{k},\bs\theta_{K}^{*})$.
\begin{theorem}
For some radius $r_d > 0$ and a triplet $(\gamma_d,\lambda_d,\mu_d)$ such that $0 \leq \gamma_d < \lambda_d \leq \mu_d$, suppose that the function $L(\bs\theta_1^{*},\bs\theta_2^{*},\dots,\bs\theta_{d-1}^{*},\bs\theta_d,\bs\theta_{d+1}^{*},\dots,\bs\theta_{K-1}^{*},\bs\theta_K^{*})$ is $\lambda_d$-strongly concave (Assumption~\ref{def:strongconN}) and $\mu_d$-smooth (Assumption~\ref{def:smoothN}), and that the GS ($\gamma_d$) condition of Assumption~\ref{def:GSN} holds. Then the population gradient AM operator $\mathcal{G}_d(\bs\theta_1,\bs\theta_2,\dots,\bs\theta_K)$ with step $\eta$ such that $0 < \eta \leq \min_{i = 1,2,\dots,K}\frac{2}{\mu_i + \lambda_i}$ is contractive over a ball $B_2(r_d,\bs\theta_d^{*})$, i.e.
\begin{equation}
\|\mathcal{G}_d(\bs\theta_1,\bs\theta_2,\dots,\bs\theta_K) - \bs\theta_d^{*}\|_2 \leq (1-\xi\eta)\|\bs\theta_d - \bs\theta_d^{*}\|_2 + \eta\gamma\sum_{\substack{i=1 \\ i\neq d}}^K\|\bs\theta_i-\bs\theta_i^{*}\|_2
\end{equation}
\vspace{-0.15in}

where $\gamma \coloneqq \max_{i=1,2,\dots,K}\gamma_i$, and $\xi \coloneqq \min_{i = 1,2,\dots,K}\frac{2\mu_i\lambda_i}{\mu_i + \lambda_i}$. \\
\vspace{0.1in}
\label{thm:contractivityG1wideN}
\end{theorem}

\subsection{Proof of Theorem~\ref{thm:contractivityG1wideN}}

\begin{eqnarray*}
&&\hspace{-0.25in}\|\mathcal{G}_d(\bs\theta_1,\bs\theta_2,\dots,\bs\theta_K) - \bs\theta_d^{*}\|_2=\|\bs\theta_d + \eta\nabla_d L(\bs\theta_1,\bs\theta_2\dots,\bs\theta_K) - \bs\theta_d^{*}\|_2\\
&&\hspace{-0.25in}\text{by the triangle inequality we further get}\\
&&\hspace{-0.25in}\leq\|\bs\theta_d + \eta\nabla_d L(\bs\theta_1^{*},\dots,\bs\theta_{d-1}^{*},\bs\theta_d,\bs\theta_{d+1}^{*},\dots,\bs\theta_K^{*}) - \bs\theta_d^{*}\|_2\\
&&\hspace{-0.25in}+ \eta\|\nabla_d L(\bs\theta_1,\dots,\bs\theta_d,\dots,\bs\theta_K)\\
&&\hspace{-0.25in}- \nabla_dL(\bs\theta_1^{*},\dots,\bs\theta_{d-1}^{*},\bs\theta_d,\bs\theta_{d+1}^{*},\dots,\bs\theta_K^{*})\|_2\\
&&\hspace{-0.25in}\text{by the contractivity of $T$ from Equation~\ref{eq:contractivityTN} from Lemma~\ref{lem:contru2N} and GS condition}\\
&&\hspace{-0.25in}\leq \left(1 - \frac{2\eta\mu_d\lambda_d}{\mu_d + \lambda_d}\right)\|\theta_d - \theta_d^{*}\|_2 + \eta\gamma_d\sum_{\substack{i=1 \\ i\neq d}}^K\|\theta_i - \theta_i^{*}\|_2.
\end{eqnarray*}

\subsection{Proof of Theorem~\ref{thm:erroruN}}

Let $\bs\theta_d^{t+1} = \Pi_d(\bs{\tilde\theta}_d^{t+1})$, where $\bs{\tilde\theta}_d^{t+1} \coloneqq \bs\theta^t_d + \eta^t\nabla_d L^1(\bs\theta_1^{t+1},\bs\theta_2^{t+1},\dots,\bs\theta_{d-1}^{t+1},\bs\theta_d^t,\bs\theta_{d+1}^{t},\dots,\bs\theta_K^t)$ ($\nabla_d L^1$ is the gradient computed with respect to a single data sample) is the update vector prior to the projection onto a ball $B_2(\frac{r_d}{2},\bs\theta_d^0)$. Let $\bs\Delta^{t+1}_d \coloneqq \bs\theta_d^{t+1} - \bs\theta_d^{*}$ and $\bs{\tilde\Delta}_d^{t+1} \coloneqq \bs{\tilde\theta}_d^{t+1} - \bs\theta_d^{*}$. Thus
\begin{eqnarray*}
     \|\bs\Delta^{t+1}_d\|_2^2 - \|\bs\Delta^t_d\|_2^2 &\leq& \|\bs{\tilde\Delta}^{t+1}_d\|_2^2 - \|\bs\Delta^t_d\|_2^2\\
    &=&    \|\bs{\tilde\theta}^{t+1}_d - \bs\theta_d^{*}\| - \|\bs\theta_d^t - \bs\theta_d^{*}\| \nonumber\\
&=&    \left<\bs{\tilde\theta}_d^{t+1}  -  \bs\theta_d^t, \bs{\tilde\theta}_d^{t+1}  +  \bs\theta_d^t  -  2\bs\theta_d^{*}\right>.
\label{eqn:tmp1N}
\end{eqnarray*}
Let $\bs{\hat{W}}_d^t \coloneqq \nabla_d L^1(\bs\theta_1^{t+1},\bs\theta_2^{t+1},\dots,\bs\theta_{d-1}^{t+1},\bs\theta_d^t,\bs\theta_{d+1}^{t},\dots,\bs\theta_K^t)$. Then we have that $\bs{\tilde\theta}_d^{t+1} - \bs\theta_d^t = \eta^t \bs{\hat{W}}_d^t$. We combine it with Equation~\ref{eqn:tmp1N} and obtain:
\begin{eqnarray*}
&&\|\bs\Delta^{t+1}_d\|_2^2 - \|\bs\Delta^t_d\|_2^2\\
&\leq& \left<\eta^t\bs{\hat{W}}_d^t,\eta^t\bs{\hat{W}}_d^t + 2(\bs\theta_d^t - \bs\theta_d^{*})\right>\\
&=& (\eta^t)^2(\bs{\hat{W}}_d^t)^{\top}\bs{\hat{W}}_d^t + 2\eta^t(\bs{\hat{W}}_d^t)^{\top}(\bs\theta_d^t - \bs\theta_d^{*})\\
&=& (\eta^t)^2\|\bs{\hat{W}}_d^t\|_2^2 + 2\eta^t\left<\bs{\hat{W}}_d^t,\bs\Delta^t_d\right>.
\end{eqnarray*}

Let $\bs W_d^t \coloneqq \nabla_d L(\bs\theta_1^{t+1},\bs\theta_2^{t+1},\dots,\bs\theta_{d-1}^{t+1},\bs\theta_d^t,\bs\theta_{d+1}^{t},\dots,\bs\theta_K^t)$. Recall that $\mathbb{E}[\bs{\hat{W}}_d^t] = \bs{W}_d^t$. By the properties of martingales, i.e. iterated expectations and tower property:
\begin{eqnarray}
\mathbb{E}[\|\bs\Delta^{t+1}_d\|_2^2] &\leq& \mathbb{E}[\|\bs\Delta^t_d\|_2^2] + (\eta^t)^2\mathbb{E}[\|\bs{\hat{W}}_d^t\|_2^2]+ 2\eta^t\mathbb{E}[\left<\bs{W}_d^t,\bs\Delta^t_d\right>]
\label{eq:edeltauN}
\end{eqnarray}
Let $\bs{W}_d^{*} \coloneqq \nabla_d L(\bs\theta_1^{*},\bs\theta_2^{*},\dots,\bs\theta_K^{*})$. By self-consistency, i.e. $\bs\theta_d^{*} = \arg\max_{\bs\theta_d\in\Omega_d}L(\bs\theta_1^{*},\dots,\bs\theta_{d-1}^{*},\bs\theta_d,\bs\theta_{d+1}^{*},\dots,\bs\theta_K^{*})$ and convexity of $\Omega_d$ we have that
\[\left<\bs W_d^{*},\bs\Delta^t_d\right> = \left<\nabla_d L(\bs\theta_1^{*},\bs\theta_2^{*},\dots,\bs\theta_K^{*}),\bs\Delta^t_d\right> \leq 0.
\]
Combining this with Equation~\ref{eq:edeltauN} we have 
\begin{eqnarray*}
\mathbb{E}[\|\bs\Delta^{t+1}_d\|_2^2] &\leq& \mathbb{E}[\|\bs\Delta^t_d\|_2^2] + (\eta^t)^2\mathbb{E}[\|\bs{\hat{W}}_d^t\|_2^2]+ 2\eta^t\mathbb{E}[\left<\bs{W}_d^t - \bs{W}_d^{*},\bs\Delta^t_d\right>].
\label{eq:edeltau2N}
\end{eqnarray*}

Define $\mathcal{G}^t_d \coloneqq \bs\theta_d^t + \eta^t\bs{W}^t_d$ and $\mathcal{G}_d^{t*} \coloneqq \bs\theta_d^{*} + \eta^t\bs{W}_d^{*}$. Thus
\begin{eqnarray*}
&&\eta^t\left<\bs W_d^t - \bs W_d^{*},\bs\Delta^t_d\right>\\
&=& \left<\mathcal{G}_d^t - \mathcal{G}_d^{t*} - (\bs\theta^t_d - \bs\theta_d^{*}), \bs\theta^t_d - \bs\theta_d^{*}\right>\\
&=& \left<\mathcal{G}_d^t - \mathcal{G}_d^{t*},\bs\theta^t_d - \bs\theta_d^{*}\right> - \|\bs\theta^t_d - \bs\theta_d^{*}\|_2^2\\
&&\hspace{-0.35in}\text{by the fact that $\mathcal{G}_d^{t*} = \bs\theta_d^{*} + \eta^t\bs W_d^{*} = \bs\theta_d^{*}$ (since $\bs W_d^{*} = 0$):}\\
&=& \left<\mathcal{G}_d^t - \bs\theta_d^{*},\bs\theta^t_d - \bs\theta_d^{*}\right> - \|\bs\theta^t_d - \bs\theta_d^{*}\|_2^2\\
&&\hspace{-0.35in}\text{by the contractivity of $\mathcal{G}^t$ from Theorem~\ref{thm:contractivityG1wideN}:}\\
&\leq& \left\{(1-\eta^t\xi)\|\bs\theta_d^t-\bs\theta_d^{*}\| + \eta^t\gamma\left(\sum_{i=1}^{d-1}\|\bs\theta_i^{t+1}-\bs\theta_i^{*}\|_2\right.\right.\left.\left.+    \sum_{i=d+1}^{K}\|\bs\theta_i^{t}-\bs\theta_i^{*}\|_2\right)\right\}\|\bs\theta^t_d - \bs\theta_d^{*}\|_2 - \|\bs\theta^t_d - \bs\theta_d^{*}\|_2^2\\
&\leq&\left\{ (1 - \eta^t\xi)\|\bs\Delta_d^t\|_2  +  \eta^t\gamma \left(\sum_{i=1}^{d-1}\|\bs\Delta_i^{t+1}\|_2  +     \sum_{i=d+1}^{K}\|\bs\Delta_i^t\|_2 \right) \right\}\cdot\|\bs\Delta^t_d\|_2  -  \|\bs\Delta^t_d\|_2^2\\
\end{eqnarray*}

Combining this result with Equation~\ref{eq:edeltau2N} gives
\begin{eqnarray*}
\mathbb{E}[\|\bs\Delta^{t+1}_d\|_2^2] &\leq& \mathbb{E}[\|\bs\Delta^t_d\|_2^2] + (\eta^t)^2\mathbb{E}[\|\bs{\hat{W}}_d^t\|_2^2]+ 2\mathbb{E}\left[ \left\{  (1 - \eta^t\xi)\|\bs\Delta_d^t\|_2 + \eta^t\gamma\left(\sum_{i=1}^{d-1}\|\bs\Delta_i^{t+1}\|_2  +     \sum_{i=d+1}^{K}\|\bs\Delta_i^t\|_2 \right)  \right\}\right.\\
&&\hspace{-0.9in}\left.\cdot\|\bs\Delta^t_d\|_2  -  \|\bs\Delta^t_d\|_2^2\right]\\
&\hspace{-1.7in}\leq&\hspace{-0.9in} \mathbb{E}[\|\bs\Delta^t_d\|_2^2] + (\eta^t)^2\sigma_{d}^2 +  2\mathbb{E}\left[ \left\{ (1 - \eta^t\xi)\|\bs\Delta_d^t\|_2  +  \eta^t\gamma\left(\sum_{i=1}^{d-1}\|\bs\Delta_i^{t+1}\|_2  +     \sum_{i=d+1}^{K}\|\bs\Delta_i^t\|_2 \right)  \right\}\right.\\
&&\hspace{-0.9in}\left.\cdot\|\bs\Delta^t_d\|_2  -  \|\bs\Delta^t_d\|_2^2 \right], \:\:\:\text{where}\\
\end{eqnarray*}
$\sigma_{d}^2 = \sup_{\substack{\bs\theta_1 \in B_2(r_1,\bs\theta_1^{*}) \\ \dots \\ \bs\theta_K \in B_2(r_K,\bs\theta_K^{*})}}\mathbb{E}[\|\nabla_d L^1(\bs\theta_1,\bs\theta_2,\dots,\bs\theta_K)\|_2^2]$.

After re-arranging the terms we obtain
\begin{eqnarray*}
&&\mathbb{E}[\|\bs\Delta^{t+1}_d\|_2^2]
\leq (\eta^t)^2\sigma_{d}^2 + (1 - 2\eta^t\xi)\mathbb{E}[\|\bs\Delta^t_d\|_2^2]+ 2\eta^t\gamma\mathbb{E} \left[ \left(\sum_{i=1}^{d-1}\|\bs\Delta_i^{t+1}\|_2  +     \sum_{i=d+1}^{K}\|\bs\Delta_i^t\|_2 \right)  \|\bs\Delta^t_d\|_2 \right]\\
&&\hspace{-0.4in}\text{apply $2ab \leq a^2 + b^2$}\\
&\leq& (\eta^t)^2\sigma_{d}^2 + (1-2\eta^t\xi)\mathbb{E}[\|\bs\Delta^t_d\|_2^2]+ \eta^t\gamma\mathbb{E} \left[\sum_{i=1}^{d-1} \left(\|\bs\Delta_i^{t+1}\|_2^2 + \|\bs\Delta^t_d\|_2^2\right) \right] + \eta^t\gamma\mathbb{E} \left[\sum_{i=d+1}^{K} \left(\|\bs\Delta_i^{t}\|_2^2 + \|\bs\Delta^t_d\|_2^2\right)  \right]\\
&=& (\eta^t)^2\sigma_{d}^2 + \mathbb{E}[\|\bs\Delta^t_d\|_2^2]\cdot\left[1 - 2\eta^t\xi  +  \eta^t\gamma(K-1) \right] + \eta^t\gamma\mathbb{E} \left[\sum_{i=1}^{d-1}\|\bs\Delta_i^{t+1}\|_2^2\right]    +  \eta^t\gamma\mathbb{E} \left[\sum_{i=d+1}^{K} \|\bs\Delta_i^{t}\|_2^2\right]\\
\end{eqnarray*}

We obtained
\begin{eqnarray*}
&&\mathbb{E}[\|\bs\Delta^{t+1}_d\|_2^2] \leq (\eta^t)^2\sigma_{d}^2 + [1-2\eta^t\xi + \eta^t\gamma(K-1)]\mathbb{E}[\|\bs\Delta^t_d\|_2^2] + \eta^t\gamma\mathbb{E} \left[\sum_{i=1}^{d-1}\|\bs\Delta_i^{t+1}\|_2^2\right] + \eta^t\gamma\mathbb{E} \left[\sum_{i=d+1}^{K} \|\bs\Delta_i^{t}\|_2^2\right]\\
&&\text{we next re-group the terms as follows}\\
&&\mathbb{E}[\|\bs\Delta^{t+1}_d\|_2^2] - \eta^t\gamma\mathbb{E} \left[\sum_{i=1}^{d-1}\|\bs\Delta_i^{t+1}\|_2^2\right]\leq [1-2\eta^t\xi + \eta^t\gamma(K-1)]\mathbb{E}[\|\bs\Delta^t_d\|_2^2]+ \eta^t\gamma\mathbb{E} \left[\sum_{i=d+1}^{K} \|\bs\Delta_i^{t}\|_2^2\right] + (\eta^t)^2\sigma_{d}^2 \\
&&\text{and then sum over $d$ from $1$ to $K$}\\
&&\mathbb{E}\left[\sum_{d=1}^K\|\bs\Delta^{t+1}_d\|_2^2\right] - \eta^t\gamma\mathbb{E} \left[\sum_{d=1}^K\sum_{i=1}^{d-1}\|\bs\Delta_i^{t+1}\|_2^2\right]\\
&\leq& [1 - 2\eta^t\xi  +  \eta^t\gamma(K - 1)]\mathbb{E}\left[\sum_{d=1}^K\|\bs\Delta^t_d\|_2^2\right]+ \eta^t\gamma\mathbb{E} \left[\sum_{d=1}^K\sum_{i=d+1}^{K} \|\bs\Delta_i^{t}\|_2^2\right]  +  (\eta^t)^2 \sum_{d=1}^K \sigma_{d}^2 \\
\end{eqnarray*}
\\

Let $\sigma = \sqrt{\sum_{d=1}^K\sigma_{d}^2}$. Also, note that
\begin{eqnarray*}
&&\hspace{-0.3in} \mathbb{E}\left[\sum_{d=1}^K\|\bs\Delta^{t+1}_d\|_2^2\right] - \eta^t\gamma(K-1)\mathbb{E} \left[\sum_{d=1}^K\|\bs\Delta_d^{t+1}\|_2^2\right]\leq \mathbb{E}\left[\sum_{d=1}^K\|\bs\Delta^{t+1}_d\|_2^2\right] - \eta^t\gamma\mathbb{E} \left[\sum_{d=1}^K\sum_{i=1}^{d-1}\|\bs\Delta_i^{t+1}\|_2^2\right]
\end{eqnarray*}
and
\begin{eqnarray*}
&&\hspace{-0.3in} [1 - 2\eta^t\xi  +  \eta^t\gamma(K - 1)]\mathbb{E}\left[\sum_{d=1}^K\|\bs\Delta^t_d\|_2^2\right]+ \eta^t\gamma\mathbb{E} \left[\sum_{d=1}^K\sum_{i=d+1}^{K} \|\bs\Delta_i^{t}\|_2^2\right]  +  (\eta^t)^2\sigma^2 \\
&\leq& [1 - 2\eta^t\xi  +  \eta^t\gamma(K - 1)]\mathbb{E}\left[\sum_{d=1}^K\|\bs\Delta^t_d\|_2^2\right]+ \eta^t\gamma(K - 1)\mathbb{E} \left[\sum_{d=1}^K \|\Delta_d^{t}\|_2^2\right]+ (\eta^t)^2\sigma^2 \\
\end{eqnarray*}

Combining these two facts with our previous results yields:
\begin{eqnarray*}
&&\hspace{-0.3in} [1-(K-1)\eta^t\gamma]\mathbb{E}\left[\sum_{d=1}^K\|\bs\Delta^{t+1}_d\|_2^2\right]\\
&\hspace{-0.2in}\leq&\hspace{-0.15in} [1 - 2\eta^t\xi  +  \eta^t\gamma(K - 1)]\mathbb{E}\left[\sum_{d=1}^K\|\bs\Delta^t_d\|_2^2\right]+ \eta^t\gamma(K - 1)\mathbb{E} \left[\sum_{d=1}^K \|\bs\Delta_d^{t}\|_2^2\right] + (\eta^t)^2\sigma^2 \\
&\hspace{-0.2in}=&\hspace{-0.15in} [1 - 2\eta^t\xi  +  2\eta^t\gamma(K - 1)]\mathbb{E}\left[\sum_{d=1}^K\|\bs\Delta^t_d\|_2^2\right]  +  (\eta^t)^2\sigma^2 \\
\end{eqnarray*}
Thus:
\begin{eqnarray*}
\mathbb{E}\left[\sum_{d=1}^K\|\bs\Delta^{t+1}_d\|_2^2\right] &\hspace{-0.1in}\leq&\hspace{-0.1in} \frac{1 - 2\eta^t\xi  +  2\eta^t\gamma(K - 1)}{1-(K-1)\eta^t\gamma}\mathbb{E}\left[\sum_{d=1}^K\|\bs\Delta^t_d\|_2^2\right]\\
&\hspace{-0.1in}+&\hspace{-0.1in} \frac{(\eta^t)^2}{1 - (K - 1)\eta^t\gamma}\sigma^2.
\end{eqnarray*}
Since $\gamma < \frac{2\xi}{3(K-1)}$, $\frac{1 - 2\eta^t\xi  +  2\eta^t\gamma(K - 1)}{1-(K-1)\eta^t\gamma} < 1$.

\subsection{Proof of Theorem~\ref{lem:errorfinal}}

To obtain the final theorem we need to expand the recursion from Theorem~\ref{thm:erroruN}. We obtained

\begin{eqnarray*}
&&\mathbb{E}\left[\sum_{d=1}^K\|\bs\Delta^{t+1}_d\|_2^2\right]\\ 
&&\leq \frac{1 - 2\eta^t[\xi  -  \gamma(K - 1)]}{1-(K-1)\eta^t\gamma}\mathbb{E}\left[\sum_{d=1}^K\|\bs\Delta^t_d\|_2^2\right]+ \frac{(\eta^t)^2}{1 - (K - 1)\eta^t\gamma}\sigma^2\\
&&= \left(1  -  \frac{\eta^t[2\xi - 3\gamma(K - 1)]}{1 - (K - 1)\eta^t\gamma}\right)\mathbb{E}\left[\sum_{d=1}^K\|\bs\Delta^t_d\|_2^2\right] + \frac{(\eta^t)^2}{1 - (K - 1)\eta^t\gamma}\sigma^2
\end{eqnarray*}
Recall that we defined $q^t$ in Theorem~\ref{thm:erroruN} as
\[q^t = 1 - \frac{1-2\eta^t\xi+2\eta^t\gamma(K-1)}{1-(K-1)\eta^t\gamma} = \frac{\eta^t[2\xi - 3\gamma(K - 1)]}{1 - (K - 1)\eta^t\gamma}
\]
and denote
\[\beta^t = \frac{(\eta^t)^2}{1 - (K-1)\eta^t\gamma}.
\]

Thus we have
\begin{eqnarray*}
&&\mathbb{E}\left[\sum_{d=1}^K\|\bs\Delta^{t+1}_d\|_2^2\right]\leq      (1-q^t)\mathbb{E}\left[\sum_{d=1}^K\|\bs\Delta^t_d\|_2^2\right]  +  \beta^t\sigma^2\\
&     \leq&      (1-q^t)\left\{(1-q^{t-1})\mathbb{E}\left[\sum_{d=1}^K\|\Delta^{t-1}_d\|_2^2\right]  +  \beta^{t-1}\sigma^2\right\}+     \beta^t\sigma^2\\
&     =&      (1 - q^t)(1 - q^{t-1})\mathbb{E}\left[\sum_{d=1}^K\|\bs\Delta^{t-1}_d\|_2^2\right]  +  (1-q^t)\beta^{t-1}\sigma^2  +\beta^t\sigma^2\\
&     \leq&      (1 - q^t)(1 - q^{t-1}) \left\{ (1 - q^{t-2})\mathbb{E} \left[\sum_{d=1}^K \|\bs\Delta^{t-2}_d\|_2^2\right]   +  \beta^{t-2}\sigma^2 \right\} + (1-q^t)\beta^{t-1}\sigma^2  +  \beta^t\sigma^2\\
&     =&      (1 - q^t)(1 - q^{t-1})(1 - q^{t-2})\mathbb{E}\left[\sum_{d=1}^K\|\bs\Delta^{t-2}_d\|_2^2\right]\\
&&    + (1 - q^t)(1 - q^{t-1})\beta^{t-2}\sigma^2   +  (1 - q^t)\beta^{t-1} \sigma^2  +  \beta^t\sigma^2\\
\end{eqnarray*}

We end-up with the following
\begin{eqnarray*}
\mathbb{E}\left[\sum_{d=1}^K\|\bs\Delta^{t+1}_d\|_2^2\right] &\leq& \mathbb{E}\left[\sum_{d=1}^K\|\bs\Delta^0_d\|_2^2\right]\prod_{i=0}^t(1 - q^i)+ \sigma^2\sum_{i=0}^{t-1}\beta^i\prod_{j=i+1}^t(1 - q^j)  +  \beta^t\sigma^2.
\end{eqnarray*}
Set $q^t = \frac{\frac{3}{2}}{t+2}$ and 
\begin{eqnarray*}
\eta^t &=& \frac{q^t}{2\xi-3\gamma(K-1) + q^t(K-1)\gamma}\\
&=& \frac{\frac{3}{2}}{[2\xi-3\gamma(K-1)](t+2) + \frac{3}{2}(K-1)\gamma}.
\end{eqnarray*}
Denote $A = 2\xi-3\gamma(K-1)$ and $B = \frac{3}{2}(K-1)\gamma$. Thus 
\[\eta^t = \frac{\frac{3}{2}}{A(t+2)+B}
\]
and
\[\beta^t = \frac{(\eta^t)^2}{1-\frac{2}{3}B\eta^t} = \frac{\frac{9}{4}}{A(t+2)[A(t+2)+B]}.
\]

\begin{eqnarray*}
&&\mathbb{E}\left[\sum_{d=1}^K\|\bs\Delta^{t+1}_d\|_2^2\right]\\ &\leq& \mathbb{E}\left[\sum_{d=1}^K\|\bs\Delta^0_d\|_2^2\right]\prod_{i=0}^t\left(1 - \frac{\frac{3}{2}}{i+2}\right)+ \sigma^2\sum_{i=0}^{t-1}\frac{\frac{9}{4}}{A(i+2)[A(i+2)+B]}\prod_{j=i+1}^t\left(1 - \frac{\frac{3}{2}}{j+2}\right)\\
&&+ \sigma^2\frac{\frac{9}{4}}{A(t+2)[A(t+2)+B]}\\
&=& \mathbb{E}\left[\sum_{d=1}^K\|\bs\Delta^0_d\|_2^2\right]\prod_{i=2}^{t+2}\left(1 - \frac{\frac{3}{2}}{i}\right)+ \sigma^2\sum_{i=2}^{t+1}\frac{\frac{9}{4}}{Ai[Ai+B]}\prod_{j=i+1}^{t+2}\left(1 - \frac{\frac{3}{2}}{j}\right)+ \sigma^2\frac{\frac{9}{4}}{A(t+2)[A(t+2)+B]}
\end{eqnarray*}

Since $A>0$ and $B>0$ thus 
\begin{eqnarray*}
&&\mathbb{E}\left[\sum_{d=1}^K\|\bs\Delta^{t+1}_d\|_2^2\right]\\ &\leq& \mathbb{E}\left[\sum_{d=1}^K\|\bs\Delta^0_d\|_2^2\right]\prod_{i=2}^{t+2}\left(1 - \frac{\frac{3}{2}}{i}\right)+ \sigma^2\sum_{i=2}^{t+1}\frac{\frac{9}{4}}{Ai[Ai+B]}\prod_{j=i+1}^{t+2}\left(1 - \frac{\frac{3}{2}}{j}\right)+ \sigma^2\frac{\frac{9}{4}}{A(t+2)[A(t+2)+B]}\\
&\leq& \mathbb{E}\left[\sum_{d=1}^K\|\bs\Delta^0_d\|_2^2\right]\prod_{i=2}^{t+2}\left(1 - \frac{\frac{3}{2}}{i}\right)+ \sigma^2\sum_{i=2}^{t+1}\frac{\frac{9}{4}}{(Ai)^2}\prod_{j=i+1}^{t+2}\left(1 - \frac{\frac{3}{2}}{j}\right)+ \sigma^2\frac{\frac{9}{4}}{[A(t+2)]^2}
\end{eqnarray*}

We can next use the fact that for any $a \in (1,2)$:
\[\prod_{i=\tau+1}^{t+2}\left(1-\frac{a}{i}\right) \leq \left(\frac{\tau+1}{t+3}\right)^a.
\]
The bound then becomes
\begin{eqnarray*}
&&\mathbb{E}\left[\sum_{d=1}^K\|\bs\Delta^{t+1}_d\|_2^2\right]\\ &     \leq&      \mathbb{E}\left[\sum_{d=1}^K\|\bs\Delta^0_d\|_2^2\right]\prod_{i=2}^{t+2}\left(1 - \frac{\frac{3}{2}}{i}\right)  + \sigma^2\sum_{i=2}^{t+1}\frac{\frac{9}{4}}{(Ai)^2}\prod_{j=i+1}^{t+2}\left(1 - \frac{\frac{3}{2}}{j}\right) + \sigma^2\frac{\frac{9}{4}}{[A(t+2)]^2}\\
&     \leq&      \mathbb{E}\left[\sum_{d=1}^K\|\bs\Delta^0_d\|_2^2\right]\left(\frac{2}{t+3}\right)^{\frac{3}{2}} + \sigma^2\sum_{i=2}^{t+1}\frac{\frac{9}{4}}{(Ai)^2}\left(\frac{i+1}{t+3}\right)^{\frac{3}{2}}  + \sigma^2\frac{\frac{9}{4}}{[A(t+2)]^2}\\
&     =&      \mathbb{E}\left[\sum_{d=1}^K\|\bs\Delta^0_d\|_2^2\right]\left(\frac{2}{t+3}\right)^{\frac{3}{2}} + \sigma^2\sum_{i=2}^{t+2}\frac{\frac{9}{4}}{(Ai)^2}\left(\frac{i+1}{t+3}\right)^{\frac{3}{2}}\\
\end{eqnarray*}

Note that $(i+1)^{\frac{3}{2}}\leq 2i$ for $i=2,3,\dots$, thus
\begin{eqnarray*}
&&\mathbb{E}\left[\sum_{d=1}^K\|\bs\Delta^{t+1}_d\|_2^2\right]\\ &     \leq&      \mathbb{E}\left[\sum_{d=1}^K\|\bs\Delta^0_d\|_2^2\right]\left(\frac{2}{t+3}\right)^{\frac{3}{2}} + \sigma^2\frac{\frac{9}{4}}{A^2(t+3)^{\frac{3}{2}}}\sum_{i=2}^{t+2}\frac{(i+1)^{\frac{3}{2}}}{i^2}\\
&     \leq&      \mathbb{E}\left[\sum_{d=1}^K\|\bs\Delta^0_d\|_2^2\right]\left(\frac{2}{t+3}\right)^{\frac{3}{2}} + \sigma^2\frac{\frac{9}{2}}{A^2(t+3)^{\frac{3}{2}}}\sum_{i=2}^{t+2}\frac{1}{i^\frac{1}{2}}\\
&&          \text{finally note that $\sum_{i=2}^{t+2}\frac{1}{i^{\frac{1}{2}}} \leq \int_{1}^{t+2}\frac{1}{x^\frac{1}{2}}dx \leq 2(t+3)^{\frac{1}{2}}$. Thus}\\
&     \leq&      \mathbb{E}\left[\sum_{d=1}^K\|\bs\Delta^0_d\|_2^2\right]\left(\frac{2}{t+3}\right)^{\frac{3}{2}} + \sigma^2\frac{9}{A^2(t+3)}\\
&&          \text{substituting $A = 2\xi-3\gamma(K-1)$ gives}\\
&     =&      \mathbb{E}\left[\sum_{d=1}^K\|\bs\Delta^0_d\|_2^2\right]\left(\frac{2}{t+3}\right)^{\frac{3}{2}} + \sigma^2\frac{9}{[2\xi-3\gamma(K-1)]^2(t+3)}\\
\end{eqnarray*}

This leads us to the final theorem.

\newpage
\section{CNNs experiments: details}
We compare SGD, Adam, and AM-Adam on the LeNet-5\cite{LeCun1998} architecture on both MNIST and Fashion-MNIST \cite{xiao2017fashion} datasets. 

Fashion-MNIST is a dataset of Zalando's article images, consisting of a training set of 60,000 examples and a test set of 10,000 examples. Each example is a 28x28 grayscale image, associated with a label from 10 classes. We intend Fashion-MNIST to serve as a direct drop-in replacement for the original MNIST dataset for benchmarking machine learning algorithms. It shares the same image size and structure of training and testing splits.

We fix the batchsize to 128, and run a hyperparameter grid search for each algorithm and dataset using the following values: weight-learning rates of 2e-M for M=2,3,4,5; batch-wise mu-increments of 1e-2,1e-5, 1e-7; epoch-wise mu-multipliers of 1, 1,1; code learning-rates of 0.1, 1 (note: only weight learning rates are varied for SGD and Adam). SGD was allowed a standard epoch-wise learning rate decay of 0.9. AM-Adam used only one subproblem iteration (both codes and weights) for each minibatch, an initial $\mu$ value of 0.01, and a maximum $\mu$ value of 1.5. In total, six total grid searches were performed.

For each hyperparameter combination, each algorithm was run on at least 5 initializations, training for 10 epochs on 5/6 of the training dataset. The mean final accuracy on the validation set (the remaining 1/6 of the training dataset) was used to select the best hyperparameters. 

\begin{figure}[!h]
	\centering
	\begin{multicols}{2}
		\includegraphics[width=0.45\textwidth]{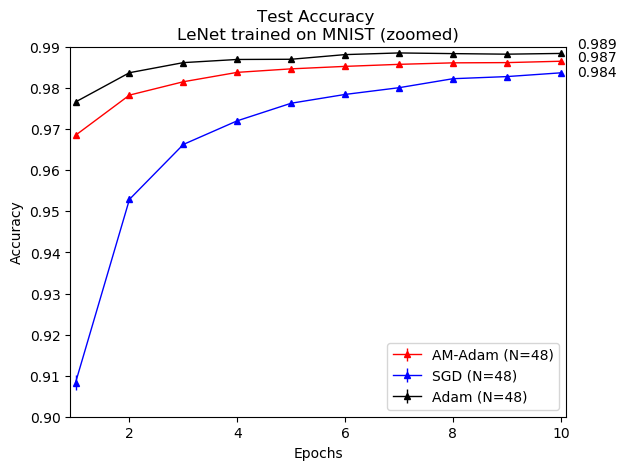} \par
		\includegraphics[width=0.45\textwidth]{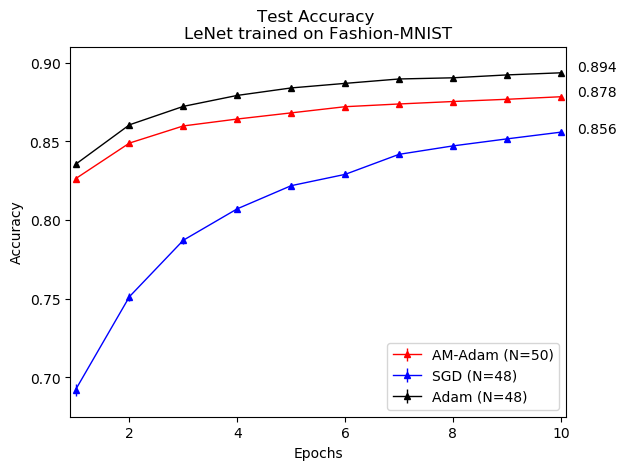}
	\vspace{-0.1in}
	\vspace{-0.15in}
	\end{multicols}
	\label{fig:CNN-LeNet-5}
\caption{CNN using LeNet-5, MNIST(left) and Fashion-MNIST (right).}
\end{figure}

Finally, each algorithm with its best hyperparameters on each dataset was used to re-train Lenet-5 with N intializations, this time evaluated on the test set. The mean performances are plotted in Figures \ref{fig:CNN-LeNet-5} for MNIST (left) and Fashion-MNIST (right).

 \vspace{0.15in}

\textbf{The winning hyperparameters for Fashion-MNIST are:}
Adam: LR=0.002
SGD: LR=0.02
AM: weight-LR= 0.002; code-LR= 1.0; batchwise $\mu$-increment=1e-5; epochwise $\mu$-multiplier=1.1

\textbf{The winning hyperparameters for MNIST are:}
Adam: LR=0.002
SGD: LR=0.02
AM: weight-LR= 0.002; code-LR= 1.0; batchwise $\mu$-increment=1e-7; epochwise $\mu$-multiplier=1.1

\newcommand{\Time}[1]{^{#1}} 
\newcommand{\bv}{\mathbf{b}}
\newcommand{\Cv}{\mathbf{C}}
\newcommand{\zv}{\mathbf{z}}
\newcommand{\Uv}{\mathbf{U}}
\newcommand{\Wv}{\mathbf{W}}
\newcommand{\Vv}{\mathbf{V}}
\newcommand{\RR}{\mathbb{R}}
\newcommand{\norm}[2]{\left\|#1\right\|_{#2}}
\section{RNN experiments: details}
\subsection{Architecture and AM Adaptation}
We also compare SGD, Adam, and AM-Adam on a standard Elman RNN architecture. That is a recurrent unit that, at time $t$, yields an output $z\Time{t}$ and hidden state $h\Time{t}$ based on a combination of input $x\Time{t}$ and the previous hidden state $h\Time{t-1}$, for $t=1,...,T$. The equations for the unit are:
\begin{align}
    h\Time{t} &= \sigma\{ \Uv x\Time{t} + \Wv h\Time{t-1} + \bv \}\\
    z\Time{t} &= \Vv h\Time{t},
\end{align}
where $\bv$ is a bias, $\sigma$ is a \textit{tanh} activation function, and $\Uv\in \RR^{d\times 1}, \Wv\in \RR^{d\times d},$ and $\Vv\in \RR^{1\times d}$ are learnable parameter matrices that do not vary with $t$. Denote with $m$ the length of one sequence element, so $x\Time{t},z\Time{t}\in\RR^{m}$. Then let $d$ be the number of hidden units, so $h\Time{t}\in\RR^d$.

We train this architecture to classify MNIST digits where each image is vectorized and fed to the RNN as a sequence of $T=784$ pixels (termed "Sequential MNIST" in \cite{le2015simple}). Thus for each $t$, the input $x\Time{t}$ is a single pixel. A final matrix $\Cv$ is then used to classify the output sequence $z\Time{t}$ using the same multinomial loss function as before:
\begin{align}
   \sum_n \mathcal{L}(y_n, ReLU(\zv_n), \Cv),
\end{align}
where $\zv_n = [z_n\Time{1}, ..., z_n\Time{784}]^{\text{T}}$ is the output sequence for the $n^{th}$ training sample, and $\Cv\in\RR^{10\times 784}$. In summary, the prediction is made only after processing all 784 pixels.

To train this family of architectures using Alt-Min, we introduce two sets of auxiliary variables (codes). First, we introduce a code for each element of the sequence just before input to the activation function:
\begin{equation}
    c\Time{t} = \Uv x\Time{t} + \Wv h\Time{t-1} + \bv
\end{equation}
where $c\Time{t}$ is the internal RNN code at time $t$. Using the "unfolded" interpretation of an RNN, we have introduced a code between each repeated "layer". Second, we treat the output sequence $\zv$ as an auxiliary variable in order to break the gradient chain between the loss function and the recurrent unit.

\subsection{Experiments}
We compare SGD, Adam, and AM-Adam on the Elman RNN architecture with hidden sizes $d=15$ and $d=50$ on the Sequential MNIST dataset. We fix the batchsize to 1024, and run a hyperparameter grid search for each algorithm using the following values: weight-learning rates of 5e-M, for M=1,2,3,4,5 (all methods); weight sparsity = 0, 0.01, 0.1 (SGD and Adam); batch-wise mu-increment 1e-M for M=2,3,4; epoch-wise mu-multiplier for 1, 1.1, 1.25, 1.5; mu-max=1, 5. SGD was allowed a standard learning-rate-decay of 0.9. AM-Adam used an initial $\mu$ value of 0.01, and used 5 subproblem iterations for both code and weight optimization subproblems.

Note: in an offline hand-tuning search, we determined that weight-sparsity only hurt Alt-Min, so it was not included in official the grid search. Also note that a larger batchsize is used for the RNN experiments because of the relatively strong dependence of the training time on batchsize. This dependence is because for each minibatch, a series of loops though $t=1,...,784$ are required.

For each hyperparameter combination, each algorithm was run on at least 3 initializations, training for 10 epochs on 5/6 of the training dataset. The mean final accuracy on the validation set (the remaining 1/6 of the training dataset) was used to select the best hyperparameters. 

Finally, each algorithm with its best hyperparameters on each dataset was used to re-train the Elman RNN with N intializations, this time evaluated on the test set. 

\textbf{The winning hyperparameters for \textit{d=15} are:}
Adam: learning rate = 0.005, L1=0;
SGD: learning rate = 0.05, L1=0;
AM-Adam: learning rate = 0.005, max-mu=1, mu-multiplier=1.1, mu-increment=0.01. Results are depicted in Figure \ref{fig:RNN15}.

\begin{figure}[!h]
    \centering
    \includegraphics[width=0.45\textwidth]{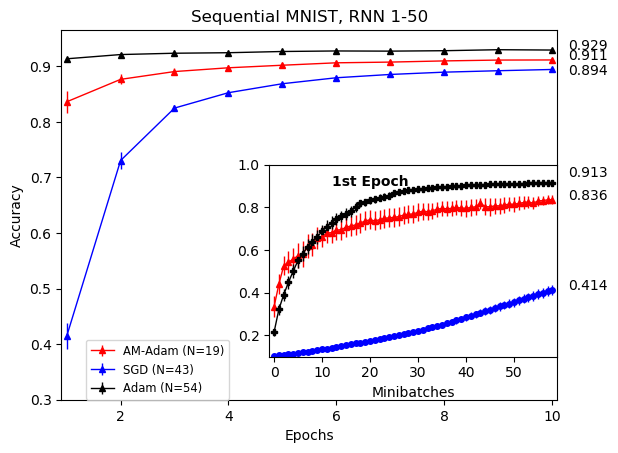}
    \vspace{-0.1in}
\caption{RNN-50, Sequential MNIST.}
  \vspace{-0.15in}
    \label{fig:RNN50}
\end{figure}

\textbf{The winning hyperparameters for \textit{d=50} are:}
Adam: learning rate = 0.005, L1=0.01;
SGD: learning rate = 0.005, L1=0;
AM-Adam: learning rate = 0.005, max-mu=1, mu-multiplier=1.0, mu-increment=0.0001. Results are depicted in Figure \ref{fig:RNN50}.

 \vspace{-0.15in}

\section{Fully connected networks: details}

Performance of the online (i.e., SGD, Adam, AM-Adam, AM-mem) and offline (i.e., AM-Adam-off, AM-mem-off, Taylor) methods are compared on the MNIST and CIFAR-10 datasets for two fully connected network architectures with two identical hidden layers of 100 and 500 units each. We also consider a different architecture with one hidden layer of 300 units for the larger HIGGS dataset. Optimal hyperparameters are reported below for each set of experiments.

\subsection{MNIST Experiments}
The standard MNIST training dataset is split into a reduced training set (first 50,000 samples) and a validation set (last 10,000 samples) for hyperparameter optimization. More specifically, an iterative bayesian optimization scheme is used to find the optimal learning rates (lr) maximizing classification accuracy on the validation set after 50 epochs of training. Rather than learning rates, for Taylor's method we optimize the $\gamma_{\textrm{prod}}$ and ${\gamma_\textrm{nonlin}}$ parameters. The procedure is repeated for five different weight initializations and for both architectures considered. Table \ref{tab:SUPPLEMENT_MNIST} reports hyperparameters yielding the highest accuracy among the 5 weight initializations.

\begin{table}[htb]
    \centering
    \begin{tabular}{ccccc}
    \toprule
        Algorithm & Hidden units per layer & $\textrm{lr}$ & $\gamma_{\textrm{prod}}$ & $\gamma_{\textrm{nonlin}}$   \\
        \midrule
         Adam & 100 & 0.0210 && \\
         Adam & 500 & 0.0005 &&\\
         SGD & 100 & 0.2030 &&\\
         SGD & 500 & 0.1497 &&\\
         AM-Adam & 100 & 0.1973 &&\\
         AM-Adam & 500 & 0.1171 &&\\
         AM-mem & 100 & 0.1737 &&\\
         AM-mem & 500 & 0.1376 &&\\
         AM-Adam-off & 100 & 0.5003 &&\\
         AM-Adam-off & 500 & 0.4834 &&\\
         AM-mem-off & 100 & 0.4664 &&\\
         AM-mem-off & 500 & 0.2503 &&\\
         Taylor & 100 && 582.8 & 54.15  \\
         Taylor & 500 && 444.2 & 111.7 \\
         \bottomrule
    \end{tabular}
    \caption{Optimal hyperparameters for fully connected networks on MNIST}
    \label{tab:SUPPLEMENT_MNIST}
    
\end{table}

\subsection{CIFAR-10 Experiments}
Similary to what done for the MNIST dataset, we split the standard CIFAR-10 training dataset into a reduced training set (first 40,000 samples) and a validation set (last 10,000 samples) used to evaluate accuracy for hyperparameter optimization. Table \ref{tab:SUPPLEMENT_CIFAR} reports hyperparameters for all the methods yielding the highest accuracy among the 5 weight initializations. Since not included in the original publication, we do not consider Taylor's method on this dataset.

\begin{table}[htbp]
    \centering
    \begin{tabular}{ccc}
    \toprule
        Algorithm & Hidden units per layer & $\textrm{lr}$  \\
        \midrule
         Adam & 100 & 0.0029 \\
         Adam & 500 & 0.0002 \\
         SGD & 100 & 0.1500 \\
         SGD & 500 & 0.1428 \\
         AM-Adam & 100 & 0.1974 \\
         AM-Adam & 500 & 0.1011 \\
         AM-mem & 100 & 0.1746 \\
         AM-mem & 500 & 0.1016 \\
         AM-Adam-off & 100 & 0.5000 \\
         AM-Adam-off & 500 & 0.4844 \\
         AM-mem-off & 100 & 0.2343 \\
         AM-mem-off & 500 & 0.2277 \\
         \bottomrule
    \end{tabular}
    \caption{Optimal hyperparameters for fully connected networks on CIFAR-10}
    \label{tab:SUPPLEMENT_CIFAR}
\end{table}

\subsection{HIGGS Experiments}
For the Higgs experiment, we compare only our best performing AM-Adam online method to Adam and SGD. Also, due to the increased computational costs associated to this dataset, we consider only one weight initialization and replace the bayesian optimization scheme with a simpler grid search. Table \ref{tab:SUPPLEMENT_HIGGS} reports the hyperparameters yielding the highest accuracy.

\begin{table}[htbp]
    \centering
    \begin{tabular}{ccc}
    \toprule
        Algorithm & Hidden units per layer & $\textrm{lr}$  \\
        \midrule
         Adam & 300 & 0.001 \\
         SGD & 300 & 0.050 \\
         AM-Adam & 300 & 0.001 \\
         \bottomrule
    \end{tabular}
    \caption{Hyperparameters used for fully connected networks on HIGGS}
    \label{tab:SUPPLEMENT_HIGGS}
\end{table}

\subsection{Related Work: ProxProp}
As we mentioned in the introduction, a closely related auxiliary methods, called ProxProp, was recently proposed in \cite{Frerix-et-al-18}. However, there are several importnant differences between ProxProp and our approach.
ProxProp only analyzes and experimentally evaluates a batch version, only briefly mentioning in section 4.2.3 that theory is extendable to mini-batch setting, without explicit convergence rates/formal proofs/experiments. Also, an assumption on eigenvalues (from eq. 14 in \cite{Frerix-et-al-18}) bounded away from zero is mentioned; however, in flat regions of optimization landscape (often found by solvers like SGD) this condition is not met, as most eigenvalues are close to zero (see, e.g. Chaudhari et al 2016). We believe that our assumptions are less restrictive from that perspective (and convergence in mini-batch setting is formally proven). Further differences include:
 (1) our formulation involves only one set of auxiliary variables/”codes” (linear z in ProxProp) rather than two (linear and nonlinear), reducing memory footprint (and potentially computing time); (2) ProxProp experiments are limited to batch mode, while we compare batch vs mini-batch vs SGD; (3) ProxProp processes both auxiliary variables and weights sequentially, layer by layer (we process auxiliary variables first, then weights in all layers independently/in parallel), which is important for ProxProp. (4) Finally, we also propose two different mini-batch methods, AM-SGD (closer to ProxProp) and AM-mem, which is very different from ProxProp as. it exploits surrogate objective method of online dictionary learning in \cite{mairal2009online}.

\begin{small}
\begin{figure*}[tbh]
\begin{multicols}{2}
\vspace{-0.2in}
    \hspace{-0.2in}
    \includegraphics[width=3.6in,height=2.6in]{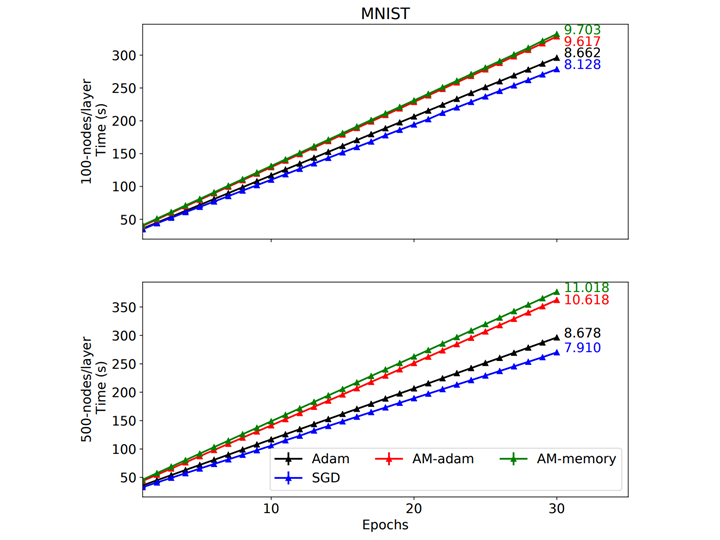}\par \vspace{-0.15in} \caption{\small Runtimes on MNIST, fully-connected architecture }\label{fig:runtimes1}
  \hspace{-0.26in}
      \includegraphics[width=3.3in,height=2.6in]{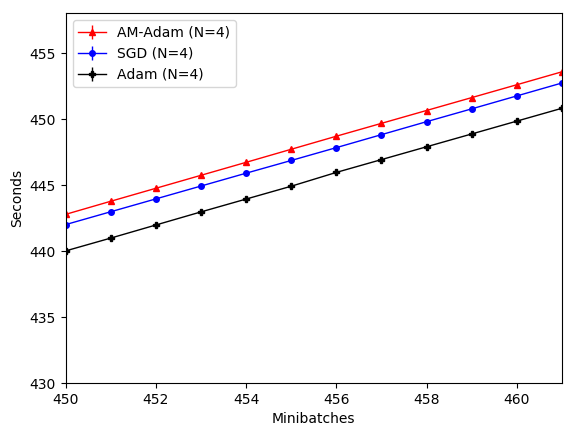}\par \vspace{-0.15in} \caption{Runtimes on MNIST, LeNet5.}\label{fig:runtimes2}
   \hspace{-0.21in}
   \end{multicols}
 \vspace{-0.25in}
\end{figure*}
\end{small}

 \subsection{Computational Efficiency: Runtimes}
 Runtime results for AM-Adam were quite comparable in most experiments to those of Adam and SGD (see Figures \ref{fig:runtimes1} and \ref{fig:runtimes2}).  Runtimes of all methods grew linearly with mini-batches/epochs, and were similar to each other: e.g., for LeNet/MNIST (Figure \ref{fig:runtimes2}), practically same slope was observed for all methods, and the runtimes were really close (e.g. 440, 442 and 443 seconds for 450 mini-batches for Adam, SGD and AM, respectively). On MNIST, using fully-connected networks (Figure \ref{fig:runtimes1}), slight increase was observed in the slope of AM versus SGD and Adam, but the times were quite comparable: e.g.,  at 30 epochs, Adam took 8.7 seconds, while  AM-SGD and AM-mem took 9.6 and 9.7 seconds, respectively. Note that we are comparing an implementation of AM which does not yet exploit parallelization; the latter is likely to provide a considerable speedup, similar to the one presented in \cite{carreira2014distributed}.

\end{document}